\documentclass{article}

\usepackage{PRIMEarxiv}

\usepackage[utf8]{inputenc} 
\usepackage[T1]{fontenc}    
\usepackage{hyperref}       
\usepackage{url}            
\usepackage{booktabs}       
\usepackage{amsfonts}       
\usepackage{nicefrac}       
\usepackage{microtype}      
\usepackage{lipsum}
\usepackage{fancyhdr}       
\usepackage{graphicx}       
\graphicspath{{media/}}     
\usepackage{multirow}
\usepackage{subcaption}
\usepackage{amsmath}
\title{Unsupervised Anomaly Prediction with N-BEATS and Graph Neural Network in Multi-variate Semiconductor Process Time Series
}

\author{
  Daniel Sorensen$^*$, Bappaditya Dey$^*$, Minjin Hwang$^*$, Sandip Halder \\
  IMEC, Kapeldreef 75, 3001 Leuven, Belgium\\
  Correspondence: \texttt{daniel.sorensen.ext@imec.be} \\ \\
  $^*$These authors contributed equally
}

\begin{document}
\maketitle

\begin{abstract}
Semiconductor manufacturing is an extremely complex and precision-driven process, characterized by thousands of interdependent parameters collected across diverse tools, process steps, and time scales. Multi-variate time-series analysis (MTSA) has emerged as a critical field for enabling real-time monitoring, fault detection, and predictive maintenance in such environments. However, applying MTSA for anomaly prediction in semiconductor fabrication presents several critical challenges. These include the high dimensionality of sensor data, severe class imbalance due to the rarity of true faults, the presence of noisy and missing measurements, and the non-stationary behaviour of production systems driven by dynamic recipe adjustments, tool ageing, and maintenance activities. Furthermore, the complex interdependencies between process variables and the delayed emergence of faults across downstream stages complicate both anomaly detection and root-cause-analysis. This paper proposes two novel approaches to advance the field from anomaly detection to anomaly prediction, an essential step toward enabling real-time process correction and proactive fault prevention. The proposed anomaly prediction framework contains two main stages: (a) training a forecasting model on a dataset assumed to contain no anomalies, and (b) performing forecast on unseen time series data. At each step, the forecast is compared with the forecast of the trained signal. Deviations beyond a predefined threshold are flagged as anomalies. The two approaches differ in the forecasting model employed. The first assumes independence between signals in the MTS and utilizes the N-BEATS model for univariate time series forecasting. The second lifts this assumption by utilizing a Graph Neural Network (GNN) to capture inter-variable relationships. Both models demonstrate strong forecasting performance up to a horizon of 20 time points and maintain stable anomaly prediction up to 50 time points. Across all tested scenarios, the GNN consistently outperforms the N-BEATS model while requiring significantly fewer trainable parameters and lower computational cost.These results position the GNN as promising solution for online anomaly forecasting to be deployed in manufacturing environments.
\end{abstract}


\section{Introduction}
\label{sec:Intro}
In recent decades, semiconductor manufacturing has steadily evolved, with node sizes shrinking from several micrometers to 5 nanometres and below in current production. As device geometries become increasingly compact and complex, the demand for precise process control, high-resolution metrology, and advanced defect inspection has intensified. These capabilities are essential to meet the stringent tolerances required in modern chip fabrication. Even minor deviations from target process parameters can result in defective components, while unexpected hardware failures—beyond the scope of predictive maintenance—can lead to unplanned tool downtime and production interruptions. Both scenarios carry significant costs on semiconductor manufacturers and their production lines and, more broadly, on any manufacturing operation. For instance, a single flow irregularity in a cleanroom environment can result in losses ranging from \$500K to \$1M per batch of scrapped wafers \cite{FlowErrorCosts}, and unplanned equipment downtime may cost between \$100K and \$2M per hour \cite{DowntimeCosts}, depending on the industry and severity. In the first scenario, process drifts and anomalies should be predicted or detected in advance and corrected in real time. In the second scenario, tool-related issues, such as sensor failures or hardware malfunction, should also be anticipated ahead of time to allow for informed decision-making. Early detection can help flag such anomalies before wafer processing begins, potentially preventing wafer loss by halting operations in advance and thus reducing production costs. However, determining how early such predictions must be made--and identifying the optimal number of time-stamp points or the appropriate time window, whether in milliseconds or seconds--remains an active area of research.

The operational state of a (semiconductor FAB) tool at any given time can be characterized by its configurable parameters (such as valve positions, nozzle settings, electrical biases, and gas flow rates) together with sensor measurements from the process chambers (e.g. pressure, temperature, and gas-species concentrations). During operation, the tool continuously records these values at fixed time-intervals, resulting in what is known as multi-variate time series (MTS) data, that captures the dynamic behaviour of the tool. This data format is fundamental to support critical tasks such as anomaly detection and prediction of process performance metrics (for example, etch rate, deposition rate, or chemical-mechanical polishing rate). Machine learning (ML) researchers have been studying MTS analysis for many years \cite{Lapedes1987}. Prior to the widespread adoption of ML techniques, traditional statistical models, such as Autoregressive Integrated Moving Average (ARIMA) \cite{ARIMA} and Autoregressive Conditional Heteroscedasticity (ARCH) \cite{ARCH}, were commonly used. However, these models assume linear dependencies and often fail to capture the complex non-linear dynamics prevalent in manufacturing process data. To address these limitations, researchers have adopted more expressive ML models, beginning with simple Multi-Layer Perceptrons (MLPs) \cite{MLP} and evolving toward advanced architectures such as deep Convolutional Neural Networks (CNNs) \cite{TCN} and Recurrent Neural Networks, particularly Long Short-Term Memory (LSTM) \cite{LSTM} networks, which are capable of learning intricate temporal and spatial patterns from data. Additionally, various auto-encoder based frameworks \cite{SAEexample} have been employed to learn compact representations of normal tool and process behaviour, aiding in feature extraction and anomaly detection.

Although statistical models and machine learning algorithms have been applied to time series analysis for over four decades, their use on multi-variate time series in the semiconductor field remains under-explored due to several key challenges such as data access limitations due to Intellectual Property constraints, data variability from variability across tools/chamber/recipes/wafers, high-dimensionality as hundreds to thousands of time-synchronized process parameters are recorded on tools. In addition to these data and modelling challenges, most work so far (as discussed in section \ref{sec:RelatedWork}) has focused on detecting anomalies offline. In this research, we present two novel deep learning based methods to address these challenges and advance the field from anomaly detection to anomaly prediction. The change to anomaly prediction allows for real time prevention of anomalies by flagging upcoming anomalies and allowing for preventive process correction. Such an achievement would allow for the prevention of both tool malfunctions as well as enhanced control of wafer processing and therefore increased production yields. Anomaly prediction however also faces the additional challenge of performing real-time forecasting on a tool, limiting the model memory and computational load to the available computing power present at the tool.

The main contributions of our research are the following:
 \begin{enumerate}
     \item \textbf{Anomaly Prediction Procedure}: We develop an anomaly prediction procedure with two main steps:
     \begin{enumerate}
         \item\textbf{Training}: A forecasting model is trained on a dataset of MTS data obtained from a semiconductor FAB tool. Although the presence of anomalies in the dataset is unknown, it is assumed that any anomalies would be scarce. This assumption enables the model to learn the non anomalous features in the MTS.
         \item\textbf{Forecasting and Anomaly Detection}: The trained model is used to forecast the MTS of a new process run. At each forecast step, the forecast is compared with that of an equivalent MTS from the training dataset. If the forecasts deviate beyond a predefined threshold, an anomaly is flagged.
     \end{enumerate}
     \item\textbf{Forecasting Model Approaches}: We implement two approaches for the forecasting model.
     \begin{enumerate}
         \item\textbf{N-BEATS}: This approach assumes variable-independence in the MTS and applies the N-BEATS model to forecast each time series separately.
         \item\textbf{Graph Neural Network}: To address the independence assumption, the second approach employs a GNN model that incorporates inter-variable correlations through a graph message passing mechanism. This structure enhances the interpretability of the tool by analyzing the final graph.
     \end{enumerate}
     \item\textbf{Unsupervised Framework}: The proposed method is completely unsupervised and utilizes minimal preprocessing. This is a significant advantage to enhance the generalizability of the framework as well decreasing the time complexity, particularly during inference. 
     \item\textbf{Performance Study}: We conducted a performance study for both approaches with various lengths of lookback and horizon windows. The study concluded that both models demonstrate strong forecasting and anomaly detection performance up to a horizon of 20 time points, with the GNN consistently outperforming the N-BEATS model while also requiring lower computational resources. 
     \item\textbf{Ablation Study}: To further explain the superior performance of the GNN, an ablation study on the graph construction procedure was performed. The key parameter in the graph construction is the \textit{top-K} which limits the maximum number of edges a node can have. Surprisingly the study demonstrated the optimal configuration was obtained when each node was limited to a single edge, effectively creating a disjoint graph with self-loops. This result raises questions on the suitability of the GNN, although its performance and low computational cost justify its continued use.
 \end{enumerate}

This demonstrates the effectiveness of our proposed method in online anomaly prediction in semiconductor process data. Additionally the proposed framework is scalable to a large number of sensors/variables and generalizable to multiple tools due to the minimal preprocessing involved and completely unsupervised training. These characteristics make the framework a promising solution for advanced process control in semiconductor manufacturing.

\section{Related Work}
\label{sec:RelatedWork}

With the growing interest in smart monitoring of manufacturing tools in the semiconductor industry, several approaches and models have been taken to predict and detect anomalies. These approaches typically fall into one of three categories \cite{ZamanzadehDarban2024}: 1) \textbf{forecasting-based} approaches predict future values of a time series, such as process or tool parameters, using a preceding window of historical sensor data. Anomalies are identified by comparing the predicted values with actual measurements, which represent known normal behaviour; significant deviations from these expected values may indicate abnormal tool behaviour or process drift; 2) \textbf{reconstruction-based} approaches also employ sliding windows to learn a low-dimensional latent space representation of normal time-series segments. During training, the model is optimized to reconstruct the original signal from this representation. At inference time, the trained model attempts to reconstruct new signals, and if it fails to do so accurately, the discrepancy is treated as an anomaly. Large reconstruction errors, when compared to the baseline of known normal patterns, indicate potential abnormalities caused by equipment faults, recipe deviations, or other process-related issues; 3) Lastly, \textbf{representation-based} approaches aim to apply models (typically, self- or semi-supervised learning techniques) to latent space representation of the time series data. The objective is to develop a robust understanding of normal patterns across processes, recipes, or sensors signals by capturing the complex temporal and contextual correlations. Anomalies in new observations are identified as deviations from this learned representation, enabling the detection of subtle or previously unseen failure modes.

In the context of multi-variate time series (MTS) anomaly detection in the semiconductor industry, several methodologies have been proposed. Notably, Liao, D. et al. \cite{Chen2020} implemented a reconstruction-based approach using a stacked autoencoder framework, deploying two autoencoders per sensor (one operating in the time domain and the other in the frequency domain) within a chemical vapour deposition tool. The model detected anomalies by observing large mean squared errors between the reconstructed signals and the actual sensor readings. 

Mellah, S. et al. \cite{Mellah2022} implemented a representation-based approach by applying Independent Component Analysis (ICA) to extract the most informative features from MTS data. These features were then used as input to decision-tree based ensemble models for anomaly detection and classification. The model was evaluated on simulated sensor data designed to resemble real production variables, with 28.6\% of the data labeled as faulty. This approach achieved an F-measure of 99.8\% for anomaly classification.

Baek, M. and Kim, S. \cite{Baek2023} transformed sliding windows of time series into a signature matrix, which was input to a Convolutional Autoencoder (CAE) in order to detect anomalies in the data. For data classified as anomalous, a residual matrix was calculated and used as input to a MLP to predict replacement segments for the anomalous parts. Finally, the KernelSHAP algorithm was employed to identify the key contributing factors behind the replacement segments. This architecture achieved classification accuracies generally above 90\% and provided a degree of interpretability regarding the causes of the anomalies. 

In the research by Hwang, R. et al. \cite{Hwang2023}, a Long Short-Term Memory Autoencoder (LSTM-AE) was combined with a Deep Support Vector Data Description (SVDD) objective function. The proposed framework includes two autoencoders: first a LSTM-AE was used to pre-train the input data and extract compact representation; then a dense layer AE was trained using a loss function derived from the SVDD objective. This SVDD-based loss encourages the latent representations of normal data to lie within a hypersphere in latent space, while anomalies are mapped outside of it. Using this approach, outliers were successfully identified in 2 out of 15 processes. Although no significant anomaly patterns were found in the remaining processes, the two flagged processes revealed instability in their corresponding chambers, as indicated by the high number of detected anomalies. Further analysis indicated that the anomalies in these chambers were caused from a similar type of malfunction.

The remaining structure of this paper is organised as follows: Section \ref{Sec:Methodology} outlines the proposed methodology, including data preprocessing, anomaly induction and model training. Section \ref{sec:Exp} details the parameter setup and practical approaches employed to train and evaluate the propsed framework. Section \ref{Sec:Results} presents the experimental results and the respective analysis. Section \ref{Sec:Limit} outlines the key limitations of the current work and explores potential directions for future research. Finally, Section \ref{Sec:Conclusion} concludes the paper. 

\section{Methodology}
\label{Sec:Methodology}
\subsection{Framework overview}\label{subsec:MethodOverview}
Before delving into the methods used at each step in the proposed framework, it is worth clarifying the goal and overall structure of the framework. The goal is to perform real-time anomaly prediction. The framework achieves this by receiving the time series of the ongoing process and forecasting the $H$ next time points into the future. This forecast is then compared to the forecast of known signals on which the model has been trained on. If the forecasts differ beyond a predefined threshold, then the anomaly is flagged. This approach works on the assumption that anomalies are rare in the training dataset and therefore the trained model will learn the non-anomalous features. When an anomalous signal is given to the model, the forecast will change dramatically and therefore the anomaly is identifiable. With this approach in mind, the following subsections will discuss the available data, anomaly induction, forecasting models and anomaly detection in further detail.
\begin{figure*}[ht]
\centerline{\includegraphics[width=0.9\textwidth]{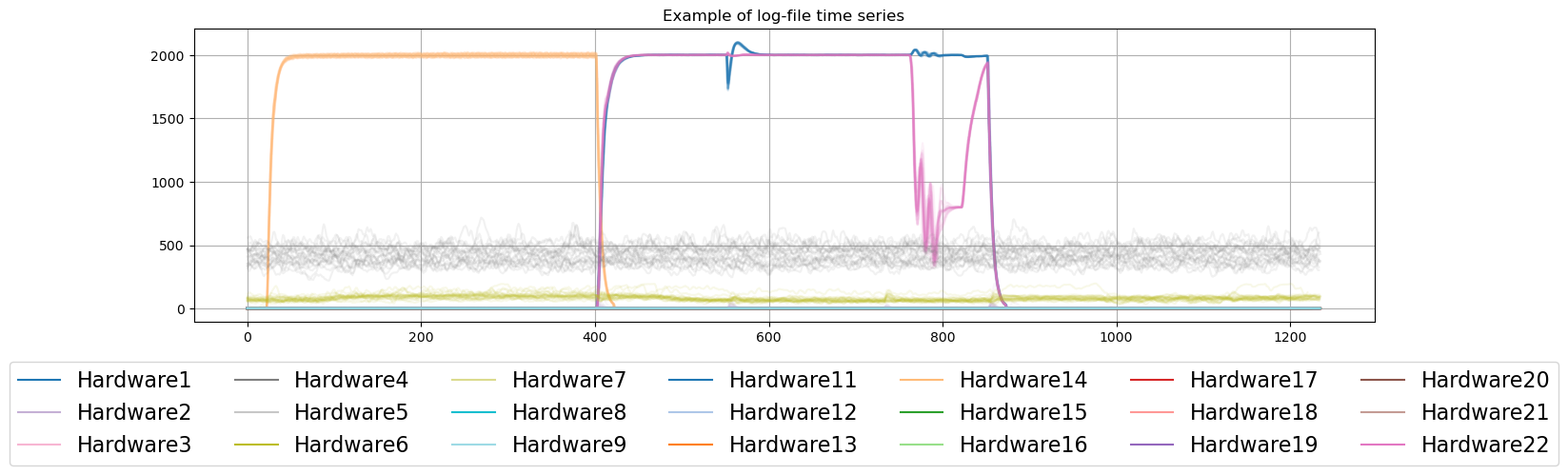}}
\caption{Example of a time series from a process in one chamber.}
\label{Fig:TimeSeries}
\end{figure*}
\subsection{Data}\label{subsec:Data}
In this study, real-world data were collected from a Coat/Develop Track tool deployed in the \textit{imec} fabrication facility. In compliance with data confidentiality requirements, the variables have been anonymized and are presented in the generalized format \textbf{\textit{HardwareXX/VariableYY}}. The collected dataset comprises the multi-variate time series of 912 process runs from 14 distinct recipes and various chambers within the tool, sampled at 0.1-second time intervals. As no ground-truth labels indicating anomalous behaviour were available, all data were treated as non-anomalous for the purpose of establishing proof-of-concept. From the 14 available recipes, this study focused on two representative cases, one containing 16 sensors and another containing 131. The process runs of both recipes had a total duration of 209.1 seconds leading to 2091 data points per time series. To illustrate the diversity of sensor behaviours, an example 125-second interval comprising 21 sensors is shown in Fig. \ref{Fig:TimeSeries}.

Based on visual inspection and domain knowledge, the time series signals can be broadly categorized into three behavioural types: 1. \textbf{Step-like}: These sensors exhibit abrupt, non-continuous transitions between discrete values. They typically correspond to binary or state-driven variables such as valve positions, fluid flow rates, or electrical biases that toggle according to recipe logic; 2. \textbf{Smooth and Noisy}: These signals display gradual transitions but are often superimposed with significant noise. They generally represent quantities not directly controlled by the recipe, such as chamber pressure, temperature, plasma density, velocity, or peak-to-peak voltage; 3. \textbf{Idle}: These sensors maintain a constant value throughout the process run, indicating that the corresponding hardware capability was not utilized in the given recipe. An example includes the flow rate of a gas not invoked by the recipe. Although idle sensors might appear redundant, excluding them would compromise the model’s ability to generalize for real-time deployment across varying recipes. Therefore, all sensors (including idle ones) are retained during model training.

\begin{figure}[htpb]
\centerline{\includegraphics[width=0.6\linewidth]{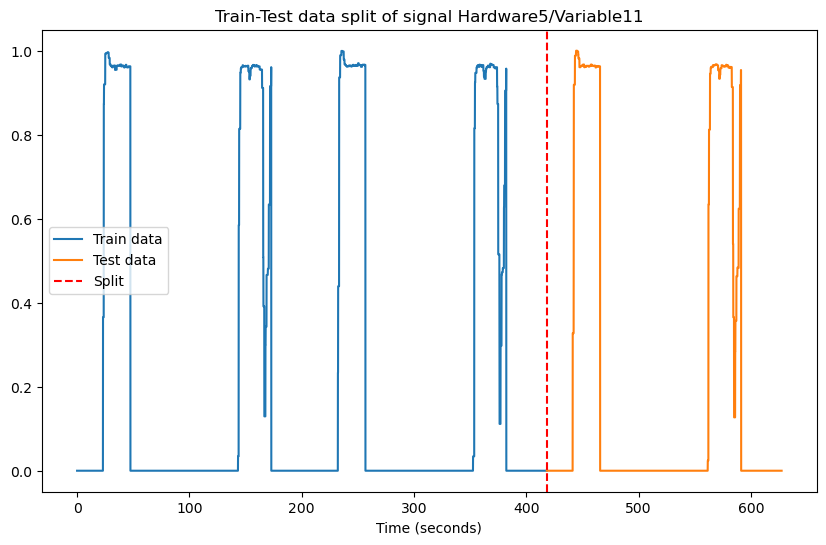}}
\caption{Train-test dataset split visualized on one time series. The training data (blue) consists of the first two thirds of the data and the test data (orange) of the last third.}
\label{fig:DataSplit}
\end{figure}

\subsection{Data preparation}\label{subsec:Dataprep}
\subsubsection{Time series processing}

To ensure the model remains lightweight and broadly applicable across different tools, preprocessing of the raw data was intentionally kept minimal. This approach reduces computational overhead and avoids introducing assumptions that may not generalize beyond the training data. As previously described, two recipes were selected for this study, each resulting in a separate dataset. The preprocessing steps outlined below were applied independently to each dataset.

Initially the data was inspected for the occurrence of missing values to assess the need for data imputation. Fortunately no missing values were found so this step was unnecessary. The only alteration of the data was therefore to apply a min-max normalization of each time series signal independently, guaranteeing the values lied within the interval [0, 1]. This ensures the numerical stability in the model computations as well as a single threshold across all variables being sufficient for anomaly detection. For real-time deployment, where the full time series is not available, min-max normalization is performed using historical values from the same recipe.

The next step is to form the training, validation and test data sets. The multi-variate time series of a process run can be represented through the matrix $\textbf{X}\in\mathbb{R}^{N\times D}$ where $N$ denotes the length of the time series and $D$ the number of variables. Row-$n$ contains the values of all variables at time instant $n$: $\textbf{X}_{n,*}:=\{{X}_{n,d}\}\,,\,for\,d\in\{1,...,D\}$\footnote{For brevity, we omit the asterisk in the row index notation (i.e. $\textbf{X}_{n}:=\textbf{X}_{n,*}$) in subsequent sections, as matrices will, unless otherwise stated, be assumed to include all variables (columns).}. Column-$d$ contains the full time series of variable $d$: $\textbf{X}_{*,d}:=\{{X}_{n,d}\}\,,\,for\,n\in\{1,...,N\}$.  
To increase the effective dataset size, three consecutive process runs were concatenated, resulting in the matrix $\textbf{X}\in\mathbb{R}^{3N\times D}$. The first two process runs were used for training the models, $\textbf{X}_{train} := \textbf{X}_{1:2N}$, where 1:2N represents the row indices {1,2,…,2N}. The remaining process run was used for testing, $\textbf{X}_{test} := \textbf{X}_{2N+1:3N}$. An example of a time series train-test split is illustrated in Fig. \ref{fig:DataSplit}.

\begin{figure*}[ht]
    \centering
    \begin{subfigure}[b]{0.3\textwidth}
        \centering
        \includegraphics[width=\textwidth]{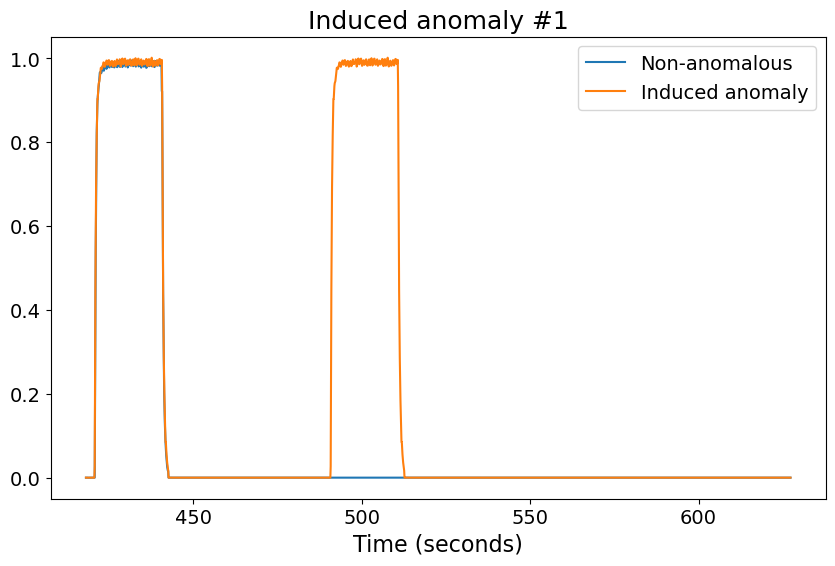}
        \caption{}
        \label{subfig:Anom1}
    \end{subfigure}
    \begin{subfigure}[b]{0.3\textwidth}
        \centering
        \includegraphics[width=\textwidth]{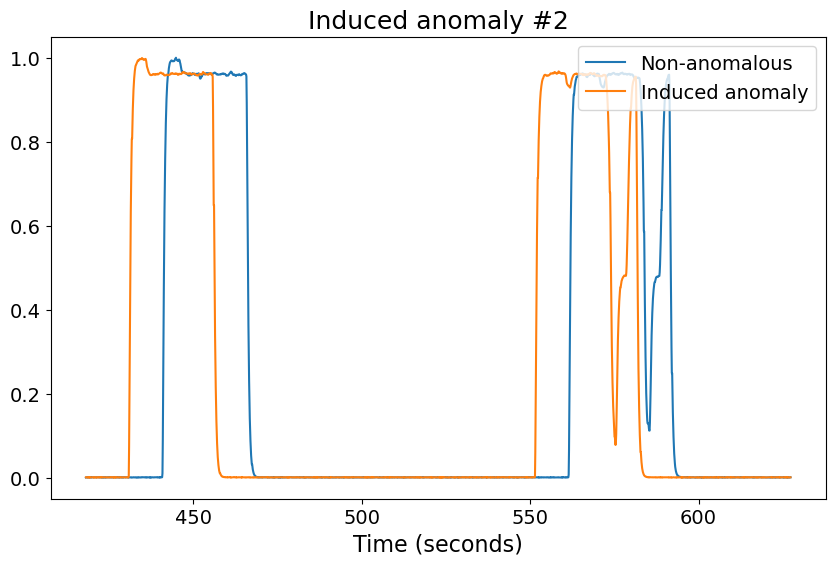}
        \caption{}
        \label{subfig:Anom2}
    \end{subfigure}
    \begin{subfigure}[b]{0.3\textwidth}
        \centering
        \includegraphics[width=\textwidth]{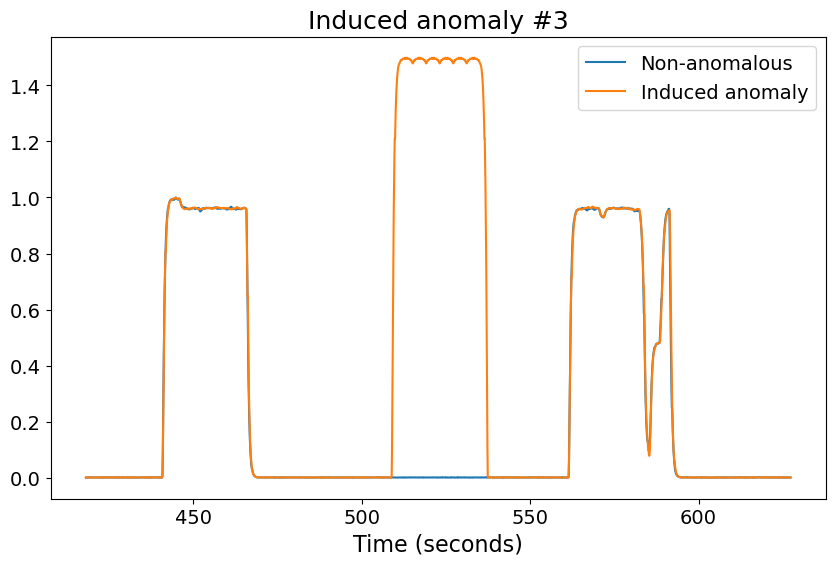}
        \caption{}
        \label{subfig:Anom3}
    \end{subfigure}
    \vspace{0.4cm}
    \begin{subfigure}[b]{0.3\textwidth}
        \centering
        \includegraphics[width=\textwidth]{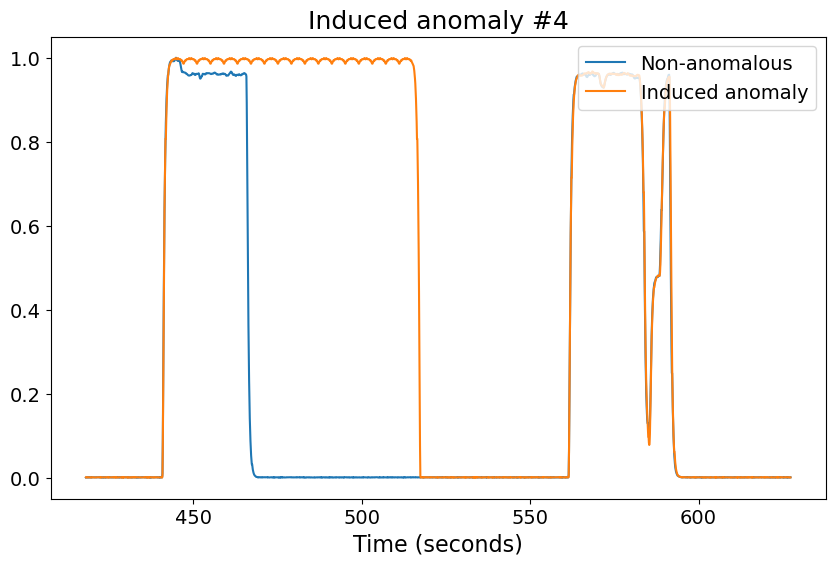}
        \caption{}
        \label{subfig:Anom4}
    \end{subfigure}
    \begin{subfigure}[b]{0.3\textwidth}
        \centering
        \includegraphics[width=\textwidth]{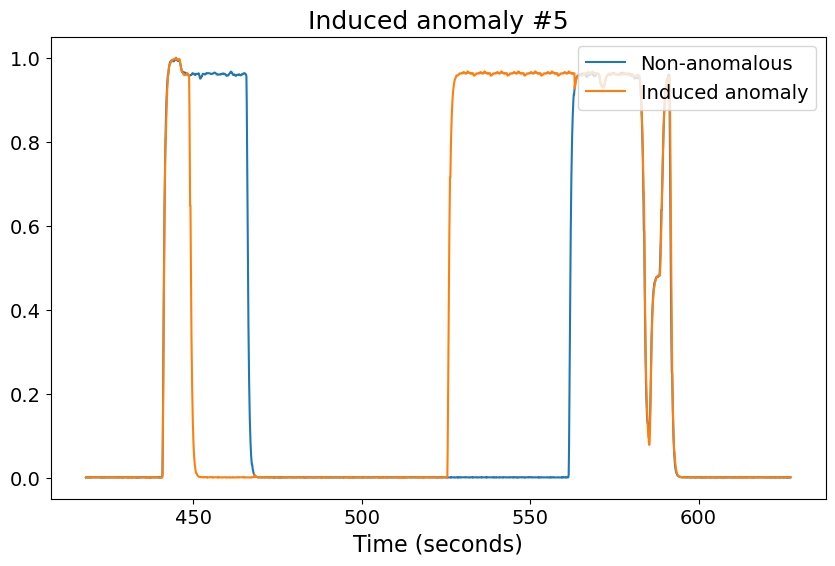}
        \caption{}
        \label{subfig:Anom5}
    \end{subfigure}
    \begin{subfigure}[b]{0.3\textwidth}
        \centering
        \includegraphics[width=\textwidth]{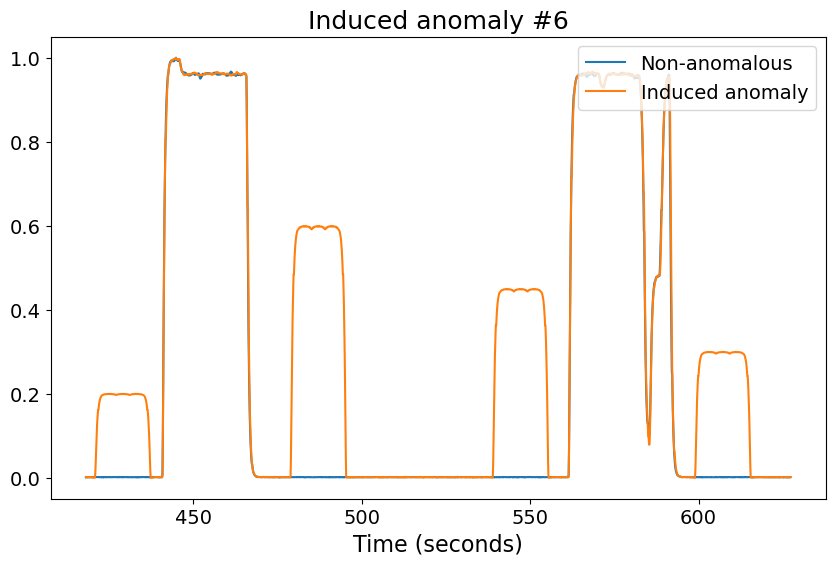}
        \caption{}
        \label{subfig:Anom6}
    \end{subfigure}
    \caption{Various examples of induced anomalies. In each example the non-anomalous signal (blue) and the corresponding induced anomaly (orange) are plotted. Figures a), c) and f) display \textit{amplitude shift} anomalies, Figure b) displays a \textit{time shift} anomaly and Figures d) and e) display \textit{step shift} anomalies.}
    \label{fig:Anomalies}
\end{figure*}

The models will be performing a forecasting task, which at a given time point $t$ can be represented by (\ref{eq:forecast}), with  $H$ denoting the horizon length, i.e. the number of points to be forecast, $L$ denoting the lookback window length, i.e. the number of past points used for the forecast, $f$ denotes the model and $\theta$ the model parameters.
\begin{equation}
    \textbf{X}_{t+1: t+H}=f\left(\textbf{X}_{t-L+1:t},\theta\right)
    \label{eq:forecast}
\end{equation}

The train data and the test data must therefore be formatted into pairs $\left(\mathcal{X}^{(t)}, \mathcal{Y}^{(t)}\right)$ := $\left(\textbf{X}_{t-L+1:t},\textbf{X}_{t+1: t+H}\right)$ to represent the input and output for each time instant $t$. In the case of the N-BEATS model, discussed in section \ref{subsec:nbeats}, it will perform the forecasting on one variable at a time, therefore the data will have the form $\left(\textbf{x}^{(t,d)}, \textbf{y}^{(t,d)}\right)$ := $\left(\textbf{X}_{t-L+1:t,d}, \textbf{X}_{t+1: t+H,d}\right)$ instead\footnote{The time (and variable) notation is moved to superscripts to allow for adequate indexation in computations to be presented in subsequent sections.}.

Finally, a validation set was created by randomly sampling 10\% of the training data. This subset was used to monitor model performance during training. 

\subsubsection{Anomaly induction}
With the preprocessing pipeline established, the data are ready for training forecasting models to learn the characteristics of non-anomalous process runs. To evaluate the models' ability to detect deviations from normal behaviour, synthetic anomalies were introduced into the test dataset. These anomalies were designed to be both significantly different from the original signals and plausible representations of real-world faults.

Anomalies were induced specifically in signals exhibiting step-like behaviour, as these are structurally simple and allow for clear visual validation. To ensure realism and diversity, three categories of anomalies were defined: 1. \textbf{Amplitude Shift} anomalies involve altering the signal amplitude within a segment. This can take the form of introducing a new step where none previously existed (see Fig. \ref{subfig:Anom1}), or modifying the amplitude of an existing step (see Fig. \ref{subfig:Anom3}). 2. \textbf{Time Shift} anomalies simulate time misalignment,  resulting in a lag between the anomalous and reference signals. The lag can be negative (early onset) or positive (delayed onset). An example of a negative time shift is shown in Fig. \ref{subfig:Anom2}. 3. \textbf{Step shift} anomalies alter the duration of a step by shifting one of its transition points in time. This results in either an extended or shortened step. Examples are illustrated in Fig. \ref{subfig:Anom4} and Fig. \ref{subfig:Anom5}.

Figure \ref{fig:Anomalies}. presents representative examples of the induced anomalies. As shown, the complexity of anomalies varies: some, like Fig. \ref{subfig:Anom1}, are highly localized and affect only a small portion of the signal, while others, such as Fig. \ref{subfig:Anom6}, contain multiple anomalies distributed across the entire signal. This diversity enables a comprehensive evaluation of the models' ability to detect a wide range of anomaly types and degrees of severity.

To quantitatively assess the models' detection capabilities, a binary label $a(t) \in {0,1}$ was assigned to each time point in the test set. A label of $a(t) = 1$ indicates the presence of an anomaly at time $t$, while $a(t) = 0$ denotes normal behaviour. These labels serve as ground truth for computing classification metrics such as precision, recall, and F1-score, once the models have predicted the anomalous time points.

\subsection{Model and training structure}
In this study, we trained and evaluated two forecasting models: N-BEATS \cite{Oreshkin2019} and a Graph Neural Network (GNN) based on the work of Ailin Deng and Bryan Hooi \cite{Deng2021}. These models were selected for their strong reported performance in time series forecasting and to enable a comparative analysis between univariate and multivariate modelling approaches. A brief overview of each model is provided in this section; for implementation details, readers are encouraged to consult the original publications.

\subsubsection{N-BEATS}\label{subsec:nbeats}
The N-BEATS model \cite{Oreshkin2019} is a deep learning architecture designed for univariate time series forecasting. It was developed to outperform traditional statistical methods while avoiding reliance on domain-specific time series components. A key feature of N-BEATS is its interpretable architecture, which allows practitioners to understand and validate the model’s forecasts.

The model is structured as a sequence of stacks, each composed of multiple blocks. Each block receives an input sequence and produces two outputs: a backcast (a reconstruction of the input) and a forecast. The first block operates on the raw input (a lookback window of length $L$), while subsequent blocks operate on the residuals —defined as the difference between the input and the cumulative backcast of previous blocks. This residual learning mechanism allows each block to iteratively refine the forecast by focusing on what previous blocks failed to capture.

Each block contains a fully connected neural network that generates two sets of coefficients, these are used to perform basis expansions for both the backcast and the forecast. The choice of basis can be:
\begin{itemize}
    \item \textit{Generic}: linear basis functions.
    \item \textit{Interpretable}: where predefined bases (e.g., polynomial for trend, trigonometric for seasonality) are used to enhance interpretability.
\end{itemize}

\begin{figure}[htpb]
    \centering
    \includegraphics[width=0.5\linewidth]{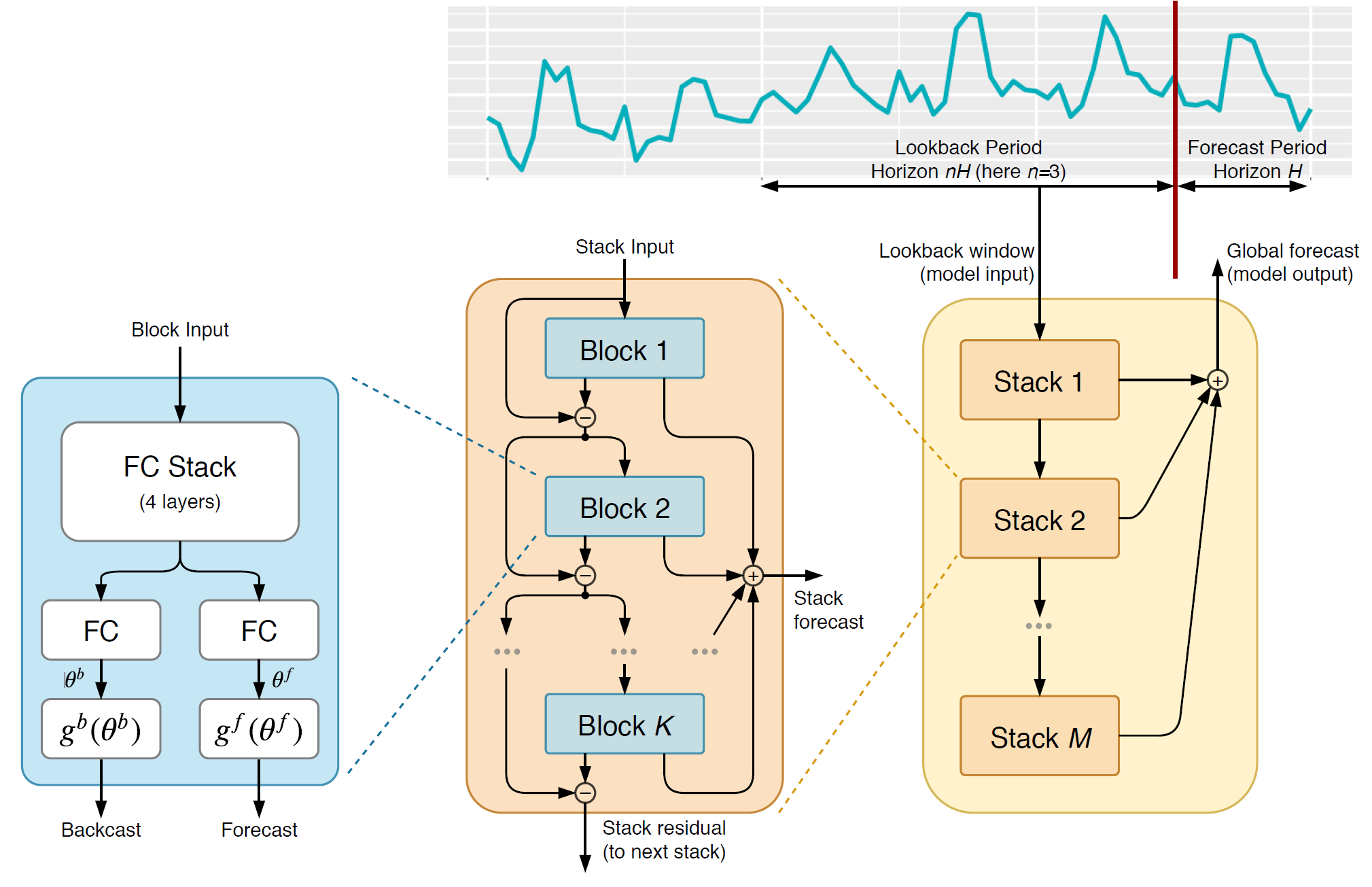}
    \caption{N-BEATS model architecture. Image taken from original article, \cite{Oreshkin2019}}
    \label{fig:nbeats}
\end{figure}

Each stack is composed of $K$ blocks and produces a partial forecast. The final forecast is obtained by summing the forecasts from all blocks. An illustration of the N-BEATS architecture, as presented in the original paper, is shown in Figure \ref{fig:nbeats}.

As the datasets studied in this work are multivariate time series, employing the N-BEATS model requires training a separate model for each variable under the assumption of independence, as no cross-variable information is utilized. This design choice significantly increases the computational load—particularly for the larger dataset, where 131 individual models must be trained to forecast all variables. The implications of this computational overhead, along with the forecasting performance of N-BEATS compared to the GNN, will be evaluated in Section \ref{Sec:Results}, enabling a discussion on the viability and efficiency of both approaches.

\subsubsection{Graph Neural Network}
\label{subsubsec:MethGNN}
The GNN implemented in this work is to a major extent the same model proposed by Ailin Deng and Bryan Hooi in \cite{Deng2021}. The difference from the model utilized in this work is the extension of the forecast horizon from one point to a defined length. This change was implemented to allow for an extended reaction time for flagged anomalies. This extension allows for an assessment of the model's performance for longer horizon forecasts as function of the backcast length and a direct comparison to the performance of the N-BEATS model.

The proposed architecture in \cite{Deng2021} contains three main components: \begin{enumerate}
    \item \textbf{Node embedding}: Used to create the graph structure, defining which variables are connected to each other.
    \item \textbf{Graph Attention Layer}: Extracts the features by combining the node embeddings and the time series information through an attention mechanism on a node and its neighbours.
    \item \textbf{Forecasting}: The extracted features are fed into fully connected layers to obtain the final forecast of the horizon.
\end{enumerate}

The node embeddings $\textbf{v}_i$ are organized into a embedding matrix $\textbf{V} \in \mathbb{R}^{D\times Emb}$, where $Emb$ denotes the embedding dimension. These embeddings are initialized randomly and trained jointly with the rest of the model parameters. Contrary to the interpretation suggested in \cite{Deng2021}, our analysis indicates that the embeddings do not incorporate any direct information from the time series data. Instead, they are optimized solely to minimize the forecasting loss. One of their primary roles is to define the graph structure via cosine similarity, as shown in Equations (\ref{eq:embed_similarity}) and (\ref{eq:Adjacency}).

\begin{equation}
    e_{i,j} = \frac{\mathbf{v}_i \cdot \mathbf{v}_j}{\Vert \mathbf{v}_i \Vert \Vert \mathbf{v}_j \Vert}
    \label{eq:embed_similarity}
\end{equation}

\begin{equation}
    A_{i,j} = 1 \quad \text{if } j \in \text{TopK}\left(\{e_{i,k}:k\in\{1,...,D\}\backslash \{i\}\,\}\right)
    \label{eq:Adjacency}
\end{equation}

Here, $\mathbf{A} \in \{0,1\}^{D\times D}$ is the binary adjacency matrix with row-$i$ indicating the edges $j$$\rightarrow$$i$, i.e. the neighbours of node-$i$. According to (\ref{eq:Adjacency}), the $K$ nodes with the highest cosine similarity to $\mathbf{v}_i$ from the neighbourhood of node-$i$. While the embeddings are also used in other parts of the model, this graph construction step is the most critical contribution. This formulation implies that the learned graph does not necessarily reflect similarity in time series behaviour. Instead, it reflects the structure that yields the best forecasting performance. While this deviates from the original interpretability claim, it enhances the model’s expressive power by allowing connections between dissimilar sensors if doing so improves predictive accuracy. However, this flexibility comes at the cost of a larger optimization space, increasing training time and complexity.

To ensure each node is connected to itself, we force self-loops in the adjacency matrix by setting $\textbf{A}_{i,i} = 1,\ i\in\{1,...,D\}$. With this adjacency matrix, the attention layer from \cite{Deng2021} can be rewritten as (\ref{eq:attention}).
\begin{equation}
    \textbf{Z}^{(t)} = ReLU\left(\left(\textbf{A}\circ \boldsymbol\alpha\right)  \mathcal{X}^{(t)}\textbf{W}^{\top}\right) 
    \label{eq:attention}
\end{equation}
Here, $\textbf{Z}^{(t)} \in \mathbb{R}^{D\times Emb}$ contains the extracted features of sample $t$, $\circ$ denotes element-wise multiplication, and $\text{W}\in \mathbb{R}^{Emb \times L}$ denotes the shared weight matrix used to embed the lookback windows of each variable. This operation aggregates the time series embeddings of connected nodes, weighted by the attention scores $\boldsymbol\alpha$. The attention weights are computed by combining time series features and  node embeddings as follows:

\begin{align}
    \textbf{g}^{(t)} &= \textbf{V} \oplus \left(\mathcal{X}^{(t)}\textbf{W}^{\top}\right) \label{eq:g} \\
    \boldsymbol\Pi_{i,j} &= \text{LeakyReLU}\left(\textbf{a} \cdot \left( \textbf{g}_i^{(t)} \oplus \textbf{g}_j^{(t)}\right)\right) \label{eq:attlogits} \\
    \boldsymbol\alpha_{i,j} &= \frac{\exp(\boldsymbol\Pi_{i,j})}{\left(\mathbf{A} \circ \exp(\boldsymbol\Pi)\right)_i \cdot\mathbf{1}_D} \label{eq:attsoftmax}
\end{align}
In these equations, $\oplus$ denotes row-wise concatenation, $\textbf{a} \in \mathbb{R}^{4\cdot Emb}$ is a learned weight vector, $exp()$ performs element-wise exponentiation and $\mathbf{1}_D$ denotes a row vector of ones with length $D$.

The final step in the GNN model is to forecast the horizon values for each variable. This is achieved using equation (\ref{eq:gnnforecast}), where the extracted features $\textbf{Z}^{(t)}$ are combined with the node embeddings $\textbf{V}$ via element-wise multiplication. The resulting matrix is then flattened into a vector and passed through a multi-layer fully connected network $f_{\theta}$ which outputs the predicted values:
\begin{equation}
    \hat{\mathcal{Y}}^{(t)} = f_{\theta} \left( \text{Flatten}\left( \mathbf{V} \circ \mathbf{Z}^{(t)} \right) \right)
    \label{eq:gnnforecast}
\end{equation}
Here, $\hat{\mathcal{Y}}^{(t)}\in\mathbb{R}^{D\cdot H}$ represents the forecasted values for all $D$ variables over a horizon of H time steps. This formulation allows the model to leverage both the learned node embeddings and the time-dependent features extracted by the attention mechanism.

\subsection{Forecast evaluation and anomaly detection}
\subsubsection{Forecast evaluation}
For both models the mean squared error (MSE) between the forecast $\hat{\mathcal{Y}}^{(t)}$ and the observed data $\mathcal{Y}^{(t)}$ is used to assess forecasting performance and is also employed as the loss function during training. In the case of the N-BEATS model, the forecast vector $\hat{\mathcal{Y}}^{(t)}$ and ground truth matrix $\mathcal{Y}^{(t)}$ are constructed by aggregating the individual forecasts $\mathbf{y}^{(t,d)}$ and $\hat{\mathbf{y}}^{(t,d)}$ across all variables. To evaluate the forecasting performance on the training dataset (comprising $2N$ time steps), the MSE is computed as:
\begin{equation}
    L_{MSE} = \frac{1}{2N - L-H+1}\sum_{t=L}^{2N-H}\left\Vert \hat{\mathcal{Y}}^{(t)} - \text{Flatten}\left(\mathcal{Y}^{(t)}\right)\right\Vert_2^2
    \label{eq:Loss}
\end{equation}

\subsubsection{Anomaly detection}
Anomalies are detected by comparing the forecast of the test dataset with the forecast of an equivalent segment from the training dataset. This approach avoids falsely flagging anomalies due to inherent forecasting errors, which may occur even in non-anomalous data. Since the training data are assumed to be anomaly-free, discrepancies between the two forecasts are more indicative of true anomalies. This will become clearer from the plots presented in section \ref{Sec:Results}.

For single-step forecasting, each time point is predicted once, and the anomaly score is computed as the average of the top-$b$ largest errors across variables, as proposed in \cite{Deng2021}. A time step is classified as anomalous if this score surpasses a defined threshold, $th$.

\begin{figure}[htpb]
    \centering
    \includegraphics[width=0.5\linewidth]{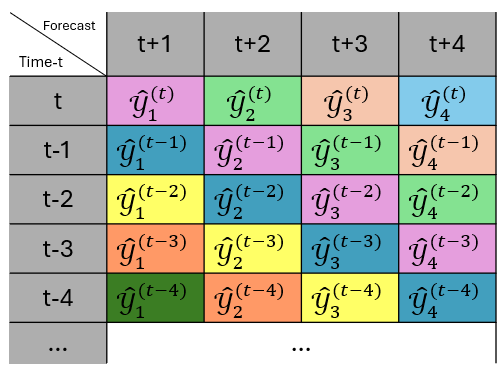}
    \caption{Forecast structure for a horizon $H$=4 of a single variable. The colour-coded diagonals highlight the forecasts of the same time point. At time step $t$, the first forecasted point is forecasted for the $H^{th}$ time. The various forecasts must be aggregated to compute an anomaly score.}
    \label{fig:forecasts}
\end{figure}

For multi-step forecasting ($H>1$), each time point is forecasted multiple times with varying lead times. As illustrated in Fig. \ref{fig:forecasts}, the colour-coded diagonals represent forecasts of the same time steps made at different points in time. To compute an anomaly score for each time step, aggregating previous forecasts can enhance the most recent forecast and improve anomaly detection. In this work, a uniform average is used to combine forecasts into a single estimate. However, future work could explore weighting schemes that prioritize forecasts made closer to the actual time step, which may be more reliable. Once a single forecast is obtained, the same procedure as for single-step forecasts is performed to detect anomalies.

\section{Experimental Setup}
\label{sec:Exp}
This section outlines the configuration used to train and evaluate the models presented in this study. We begin by describing the datasets employed, followed by the training conditions and hyperparameters specific to each model. Finally, we detail the software and hardware infrastructure used for implementation.

\subsection{Datasets}
To address the first key aspect, various anomalies were induced in the two collected datasets as described in sections \ref{subsec:Data} and \ref{subsec:Dataprep}. By denoting the original dataset with 131 variables as \textit{Dataset-A} and the dataset with 16 variables as \textit{Dataset-B}, the datasets with induced anomalies are described in Table \ref{tab:datasets}. A total of 14 datasets were created , 9 datasets from \textit{Dataset-A} and 5 from  \textit{Dataset-B}. The datasets vary in terms of the number of signals which were attacked, i.e. suffered shifts, the category of the induced anomaly and the ratio of the initial signal which was altered. These datasets create a diverse set of conditions with which to evaluate the models.

\begin{table}[]
\centering
\resizebox{0.8\linewidth}{!}{\begin{tabular}{c|c|c|c|c|}
Dataset & \#Variables          & \begin{tabular}[c]{@{}c@{}}\#Attacked \\ Variables\end{tabular} & \begin{tabular}[c]{@{}c@{}}Anomaly \\ category\end{tabular} & Anomaly ratio \\ \hline
A-1     & \multirow{9}{*}{131} & 1                                                               & Amplitude shift                                             & 33\%                                                                                              \\ \cline{1-1} \cline{3-5} 
A-2     &                      & 1                                                               & Step shift                                                  & 29\%                                                                                              \\ \cline{1-1} \cline{3-5} 
A-3     &                      & 1                                                               & Step shift                                                  & 35\%                                                                                              \\ \cline{1-1} \cline{3-5} 
A-4     &                      & 2                                                               & Amplitude shift                                             & 11\%                                                                                              \\ \cline{1-1} \cline{3-5} 
A-5     &                      & 1                                                               & Amplitude shift                                             & 15\%                                                                                              \\ \cline{1-1} \cline{3-5} 
A-6     &                      & 1                                                               & Time shift                                                  & 29\%                                                                                              \\ \cline{1-1} \cline{3-5} 
A-7     &                      & 1                                                               & Time + Amplitude shifts                                     & 43\%                                                                                              \\ \cline{1-1} \cline{3-5} 
A-8     &                      & 1                                                               & Time + Step shifts                                          & 51\%                                                                                              \\ \cline{1-1} \cline{3-5} 
A-9     &                      & 1                                                               & Time shift                                                  & 29\%                                                                                              \\ \hline
B-1     & \multirow{5}{*}{16}                     & 1                                                               & Amplitude shift                                             & 10\%                                                                                              \\ \cline{1-1} \cline{3-5} 
B-2     &                      & 1                                                               & Step shift                                                  & 9\%                                                                                               \\ \cline{1-1}\cline{3-5} 
B-3     &                      & 2                                                               & Step + Amplitude shifts                                     & 47\%                                                                                              \\ \cline{1-1}\cline{3-5} 
B-4     &                      & 2                                                               & Step + Amplitude shifts                                     & 45\%                                                                                              \\ \cline{1-1}\cline{3-5} 
B-5     &                      & 1                                                               & Amplitude shift                                             & 11\%                                                                                              \\ \hline
\end{tabular}}
\caption{Information on datasets used for model evaluation. Each dataset differs in terms of the anomaly induced regarding  number of attacked signals, anomaly category and ratio the original signal which was altered.}
\label{tab:datasets}
\end{table}

\subsection{Hyperparameters and training method}
Each model was trained using a distinct set of hyperparameters, which are summarized in Table \ref{tab:hyperp}. With the exception of \textit{top-K} all parameters were kept fixed. As for the \textit{top-K} parameter, an ablation study was performed on it's impact on the GNN performance. 

The training procedure was consistent across models: a maximum of 1000 epochs was allowed, with early stopping triggered if the validation loss did not improve for 100 consecutive epochs. The batch size was set to 32, and we used the Adam optimizer \cite{adam}, with an initial learning rate of 0.001. The learning rate was reduced by a factor of 0.5 if the validation loss failed to improve over 5 epochs.

To evaluate model performance across different temporal scales, the following lookback–horizon pairs were used: (10, 3), (20, 5), (50, 10), (100, 20), (200, 50), and (500, 100). For anomaly detection, the maximum error across all variables was used to score each time point (i.e., top-$b$ = 1), and a fixed threshold of $th$ = 0.1 was applied to classify anomalies.

\begin{table}[]
\centering
\resizebox{0.6\linewidth}{!}{\begin{tabular}{ccccc}
\multicolumn{2}{c}{N-BEATS}                                              &                       & \multicolumn{2}{c}{GNN}                                               \\ \cline{1-2} \cline{4-5} 
\multicolumn{1}{|c|}{\#stacks}            & \multicolumn{1}{c|}{2}       & \multicolumn{1}{c|}{} & \multicolumn{1}{c|}{Emb}                   & \multicolumn{1}{c|}{128} \\ \cline{1-2} \cline{4-5} 
\multicolumn{1}{|c|}{\#blocks per stack}  & \multicolumn{1}{c|}{2}       & \multicolumn{1}{c|}{} & \multicolumn{1}{c|}{TopK}                  & \multicolumn{1}{c|}{\{1, 3, 6, 9, 12, 15\}}   \\ \cline{1-2} \cline{4-5} 
\multicolumn{1}{|c|}{\#basis functions}   & \multicolumn{1}{c|}{4}       & \multicolumn{1}{c|}{} & \multicolumn{1}{c|}{$f_{\theta}$ \#layers} & \multicolumn{1}{c|}{1}   \\ \cline{1-2} \cline{4-5} 
\multicolumn{1}{|c|}{Basis}               & \multicolumn{1}{c|}{Generic} &                       &                                            &                          \\ \cline{1-2}
\multicolumn{1}{|c|}{hidden layers units} & \multicolumn{1}{c|}{128}     &                       &                                            &                          \\ \cline{1-2}
\end{tabular}}
\caption{Hyperparameters for the model architectures of the N-BEATS and the GNN models.}
\label{tab:hyperp}
\end{table}

\subsection{Software and Hardware}
The implementations of both models were obtained from publicly available repositories: N-BEATS from \cite{NBeatsPRemy}, and the Graph Neural Network (GNN) via the authors’ official implementation referenced in \cite{Deng2021}. All training and inference were conducted on a server equipped with an Intel(R) Xeon(R) Gold 6134 CPU @ 3.20GHz (8 cores) and two NVIDIA Tesla V100 PCIe GPUs.

\section{Results \& Discussion}
\label{Sec:Results}
This section presents the empirical results obtained from the experiments described above. We first report the best performance metrics achieved by each method, followed by an ablation study analyzing the impact of the \textit{top-K} parameter on GNN performance. Lastly, we end the section with an analysis of the computational complexity of the two models.

\subsection{Performance}
\label{subsec:ResPerf}
\begin{figure}[htbp]
    \centering
    \includegraphics[width=0.7\linewidth]{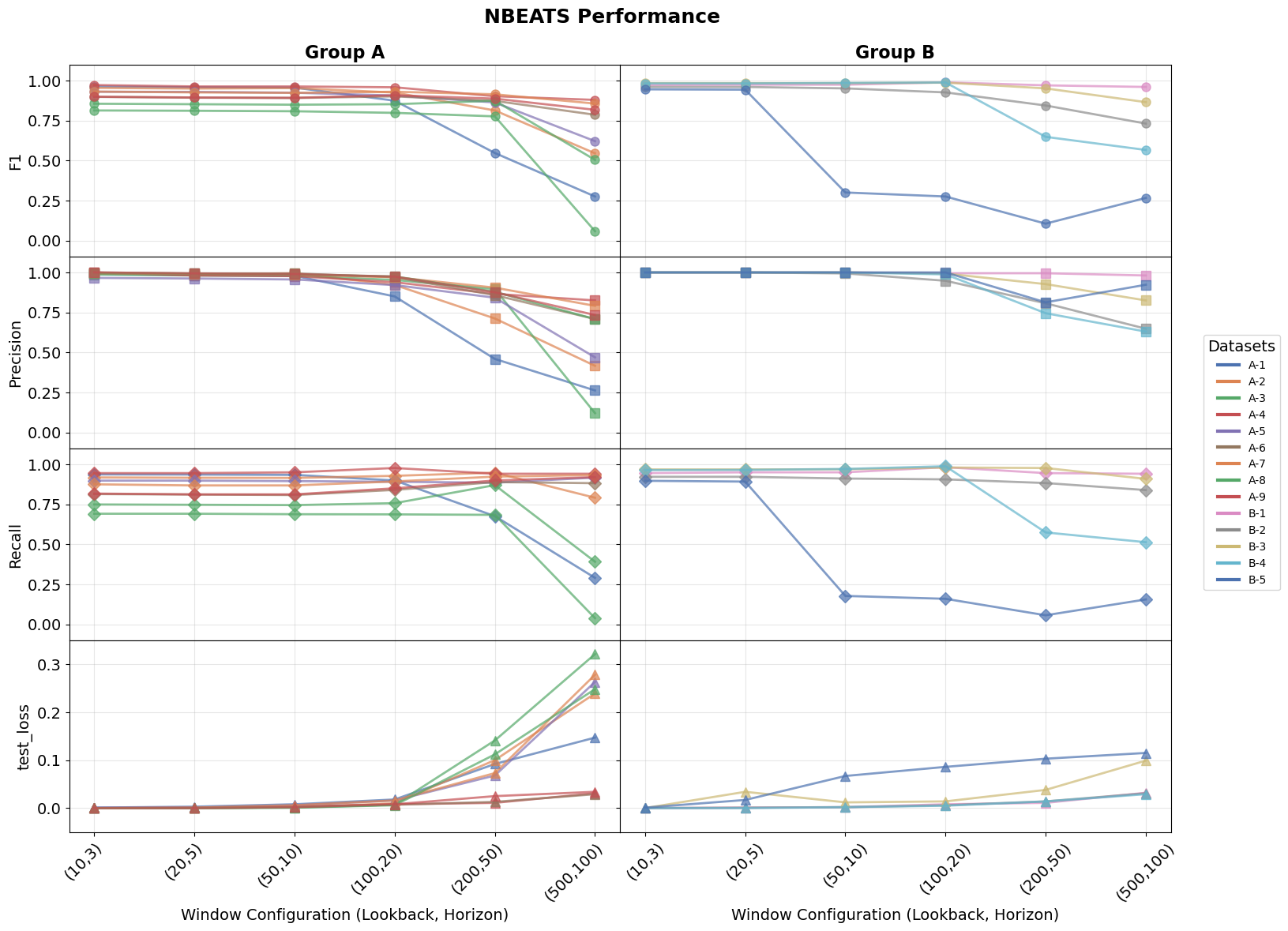}
    \caption{N-BEATS model performance: F1-Score, Precision, Recall, and test loss (MSE) across training window configurations.}
    \label{fig:nbeats_result}
\end{figure}

\begin{figure}[htpb]
    \centering
    \includegraphics[width=0.7\linewidth]{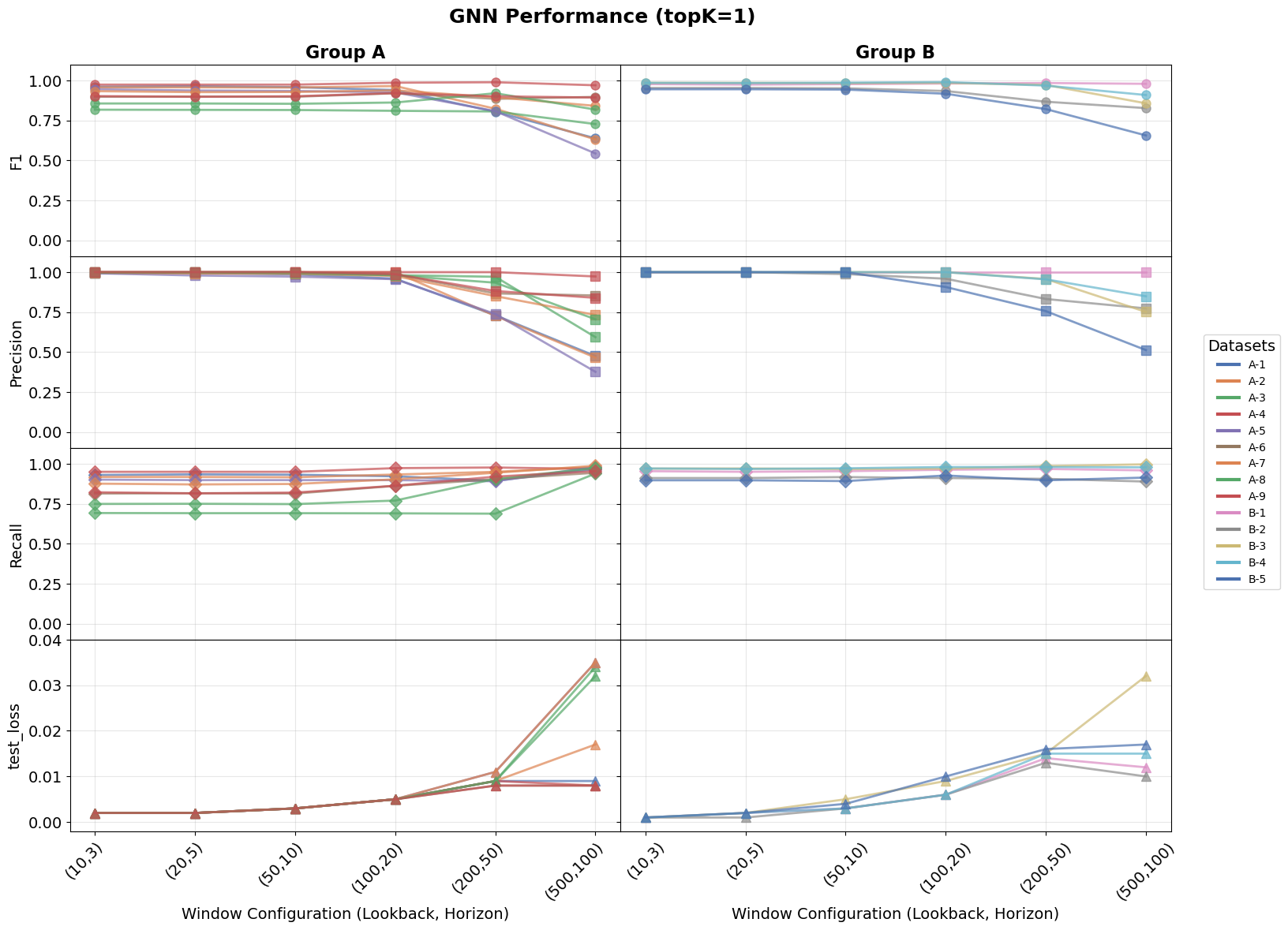}
    \caption{GNN model performance (\textit{top-K}=1): F1-Score, Precision, Recall, and test loss (MSE) across training window configurations.}
    \label{fig:gnn_result}
\end{figure}

Figures \ref{fig:nbeats_result}. and \ref{fig:gnn_result}. display the performance of the two models in terms of F1-score, Precision and Recall for the classification of anomalous time points as well as the test loss of the model's forecasts. A performance result is obtained for each window (lookback, horizon) configuration and the results are separated according to the dataset group (A or B). Only one architecture of the N-BEATS was implemented, whereas the GNN was implemented for various values of \textit{top-K}. As will be presented in section \ref{subsec:topk}, the best results were obtained for \textit{top-K} = 1, which are therefore the results presented for comparison with the N-BEATS.


An examination of the performance results reveals several consistent trends across both models. A primary observation is the decline in performance as the horizon window increases. This behaviour is expected, as forecasting further into the future inherently introduces greater uncertainty and complexity. Despite this challenge, both models maintain an F1-score above 75\%  up to 100 point long horizons, indicating a generally effective identification of anomalous time points.

A closer comparison of Precision and Recall metrics shows that Recall values are consistently lower than their corresponding Precision values. This suggests that while the models are proficient at identifying true anomalies, they also tend to flag additional time points as anomalous. In the context of semiconductor manufacturing, this trade-off is particularly relevant: ensuring the detection of all true anomalies is critical to prevent defective wafers from proceeding through the production pipeline. However, excessive false positives may lead to unnecessary process interruptions or corrective actions, which are undesirable from an operational standpoint.

Finally, an analysis of the test loss values reveals that for the first three horizon configurations, both models perform comparably. However, as the horizon window increases, the GNN model demonstrates a clear advantage, achieving loss values that are an order of magnitude lower than those of the N-BEATS model. This superior forecasting capability is also reflected in the slower degradation of other performance metrics with increasing horizon size, further underscoring the robustness of the GNN approach in long-term anomaly detection.

\begin{figure*}[ht]
    \centering
    \begin{subfigure}[b]{0.32\textwidth}
        \centering
        \includegraphics[width=\textwidth]{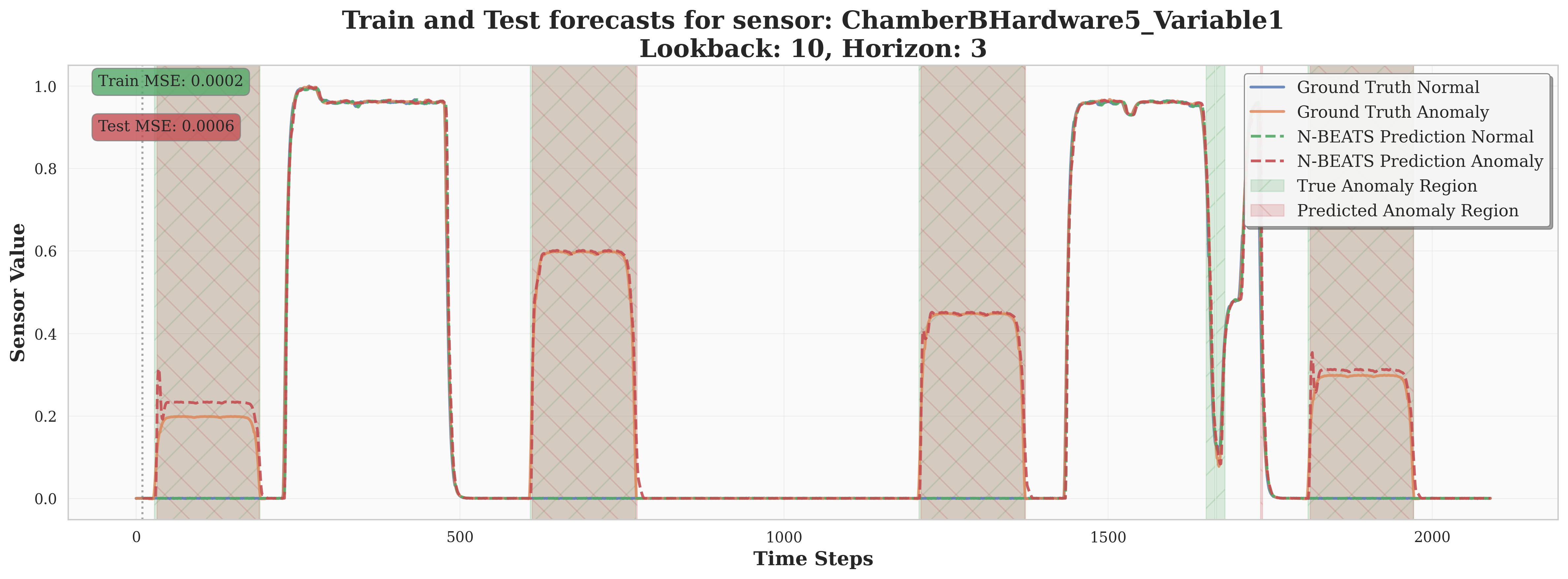}
    \end{subfigure}
    \begin{subfigure}[b]{0.32\textwidth}
        \centering
        \includegraphics[width=\textwidth]{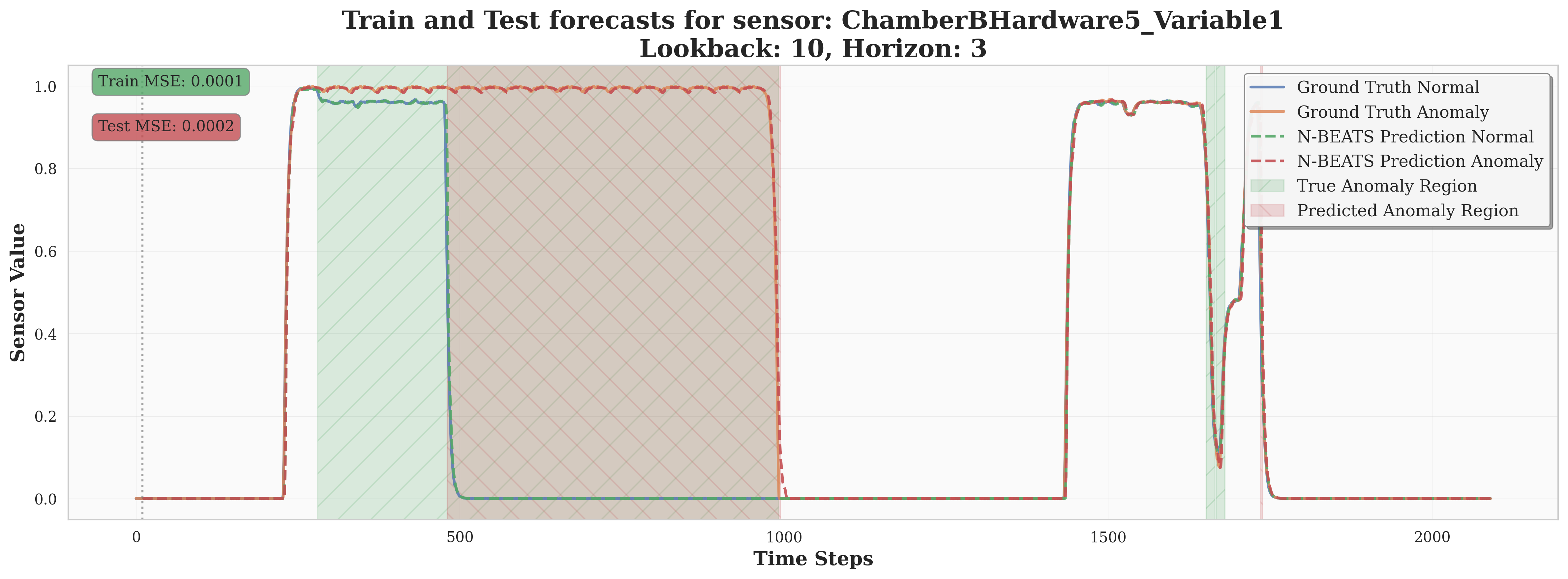}
    \end{subfigure}
    \begin{subfigure}[b]{0.32\textwidth}
        \centering
        \includegraphics[width=\textwidth]{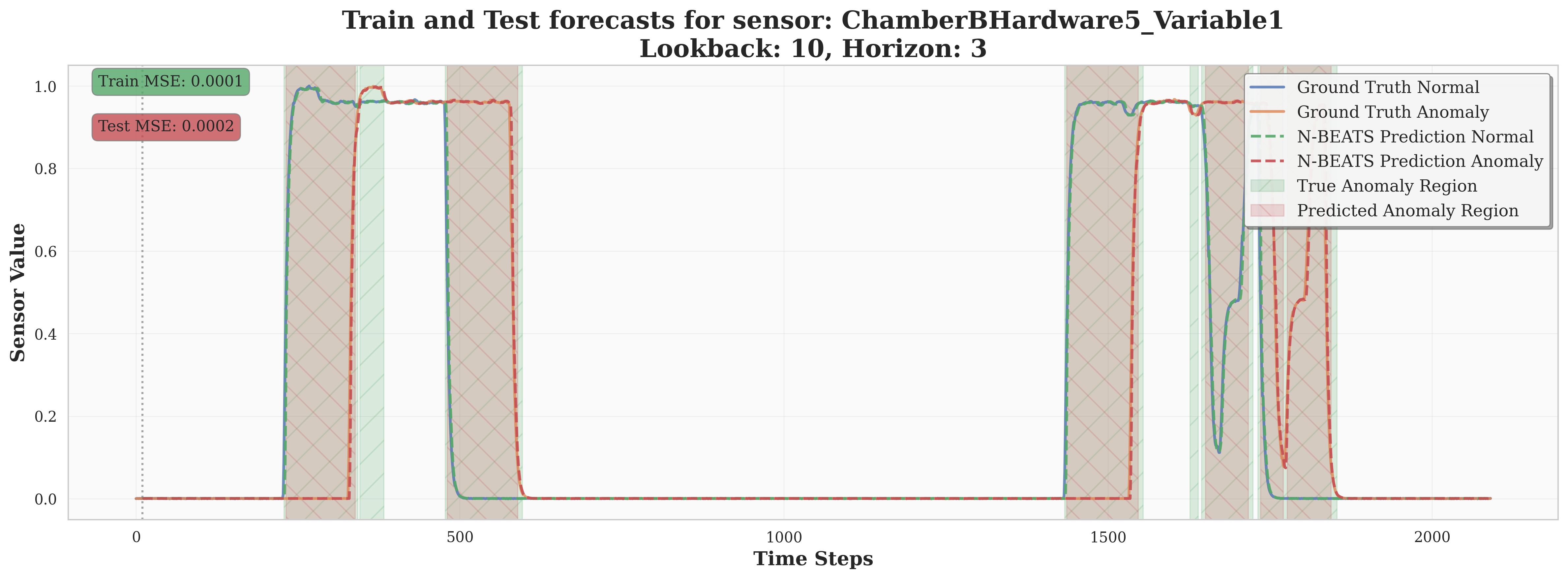}
    \end{subfigure}
    \vspace{0.4cm}
    \begin{subfigure}[b]{0.32\textwidth}
        \centering
        \includegraphics[width=\textwidth]{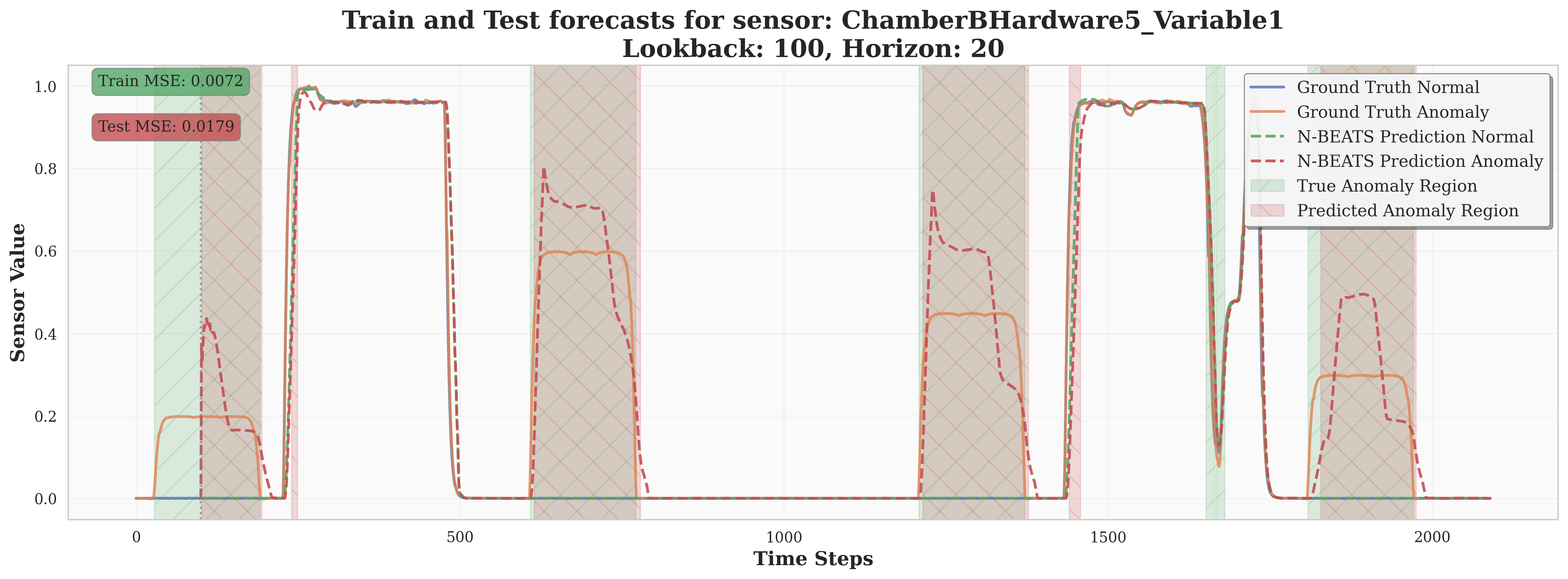}
        \caption{N-BEATS forecasts for dataset A-1}
        \label{subfig:nb4}
    \end{subfigure}
    \begin{subfigure}[b]{0.32\textwidth}
        \centering
        \includegraphics[width=\textwidth]{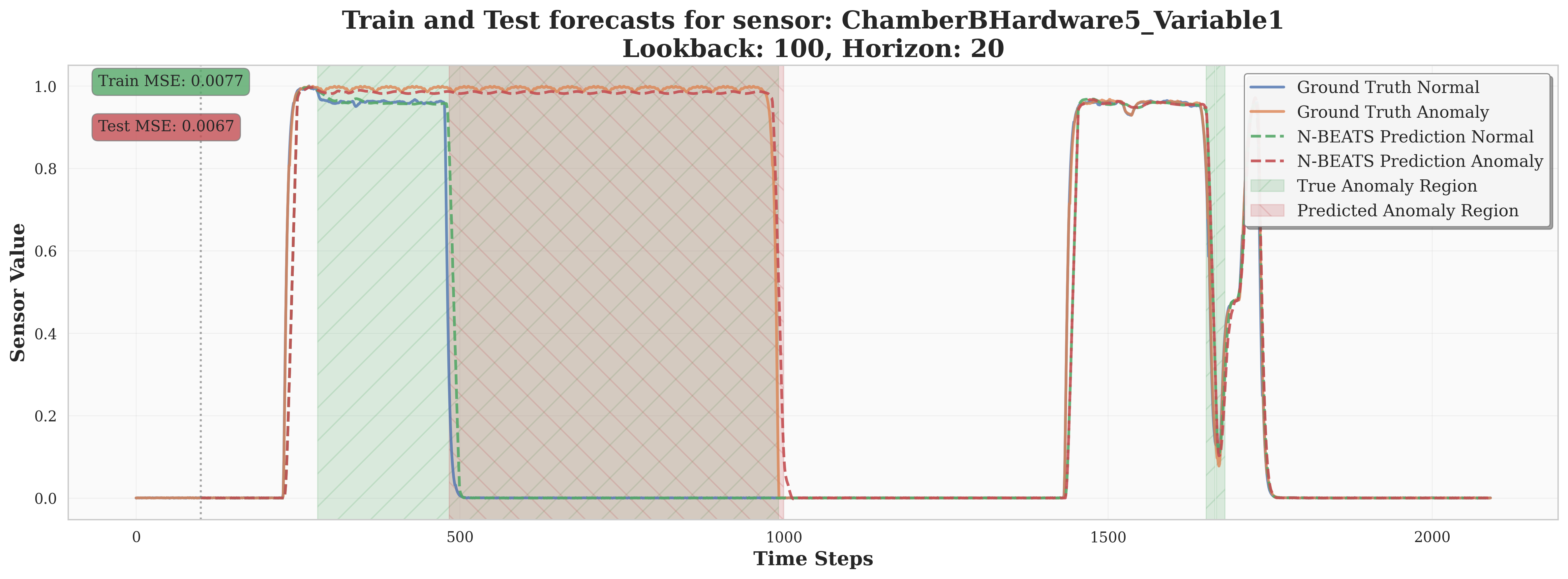}
        \caption{N-BEATS forecasts for dataset A-3}
        \label{subfig:nb5}
    \end{subfigure}
    \begin{subfigure}[b]{0.32\textwidth}
        \centering
        \includegraphics[width=\textwidth]{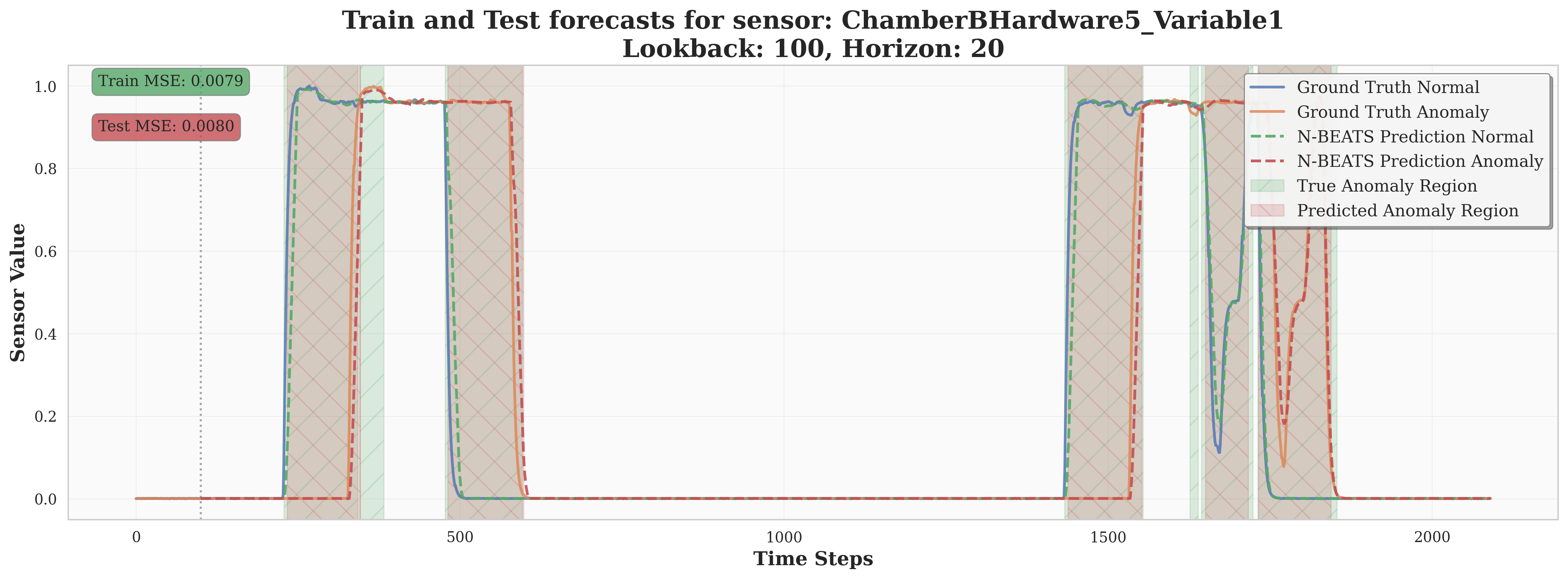}
        \caption{N-BEATS forecasts for dataset A-9}
        \label{subfig:nb6}
    \end{subfigure}
    \vspace{0.4cm}
    \begin{subfigure}[b]{0.32\textwidth}
        \centering
        \includegraphics[width=\textwidth]{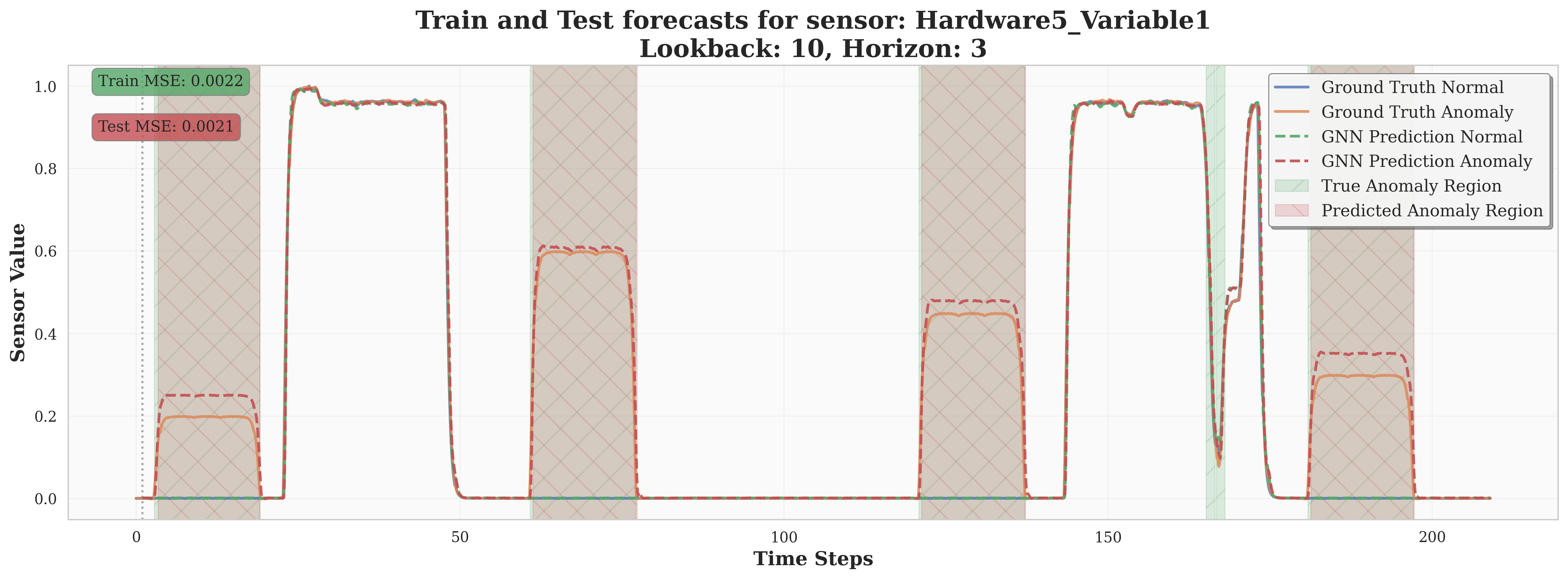}
    \end{subfigure}
    \begin{subfigure}[b]{0.32\textwidth}
        \centering
        \includegraphics[width=\textwidth]{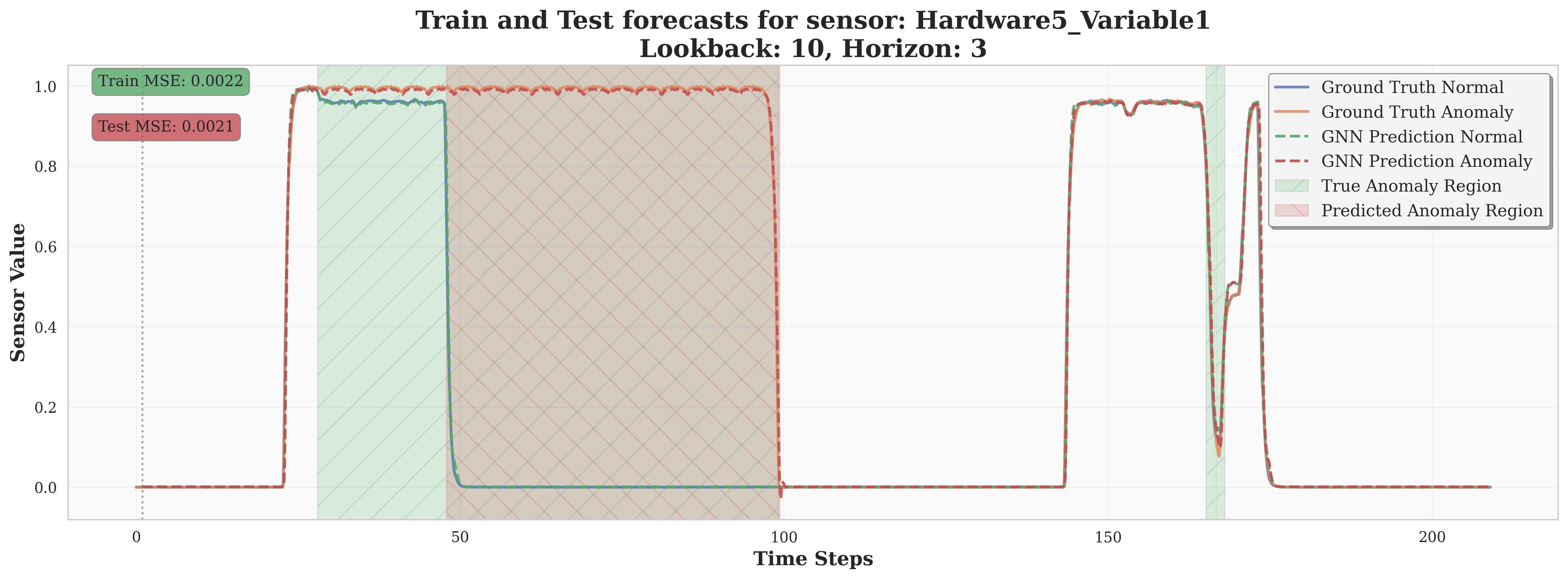}
    \end{subfigure}
    \begin{subfigure}[b]{0.32\textwidth}
        \centering
        \includegraphics[width=\textwidth]{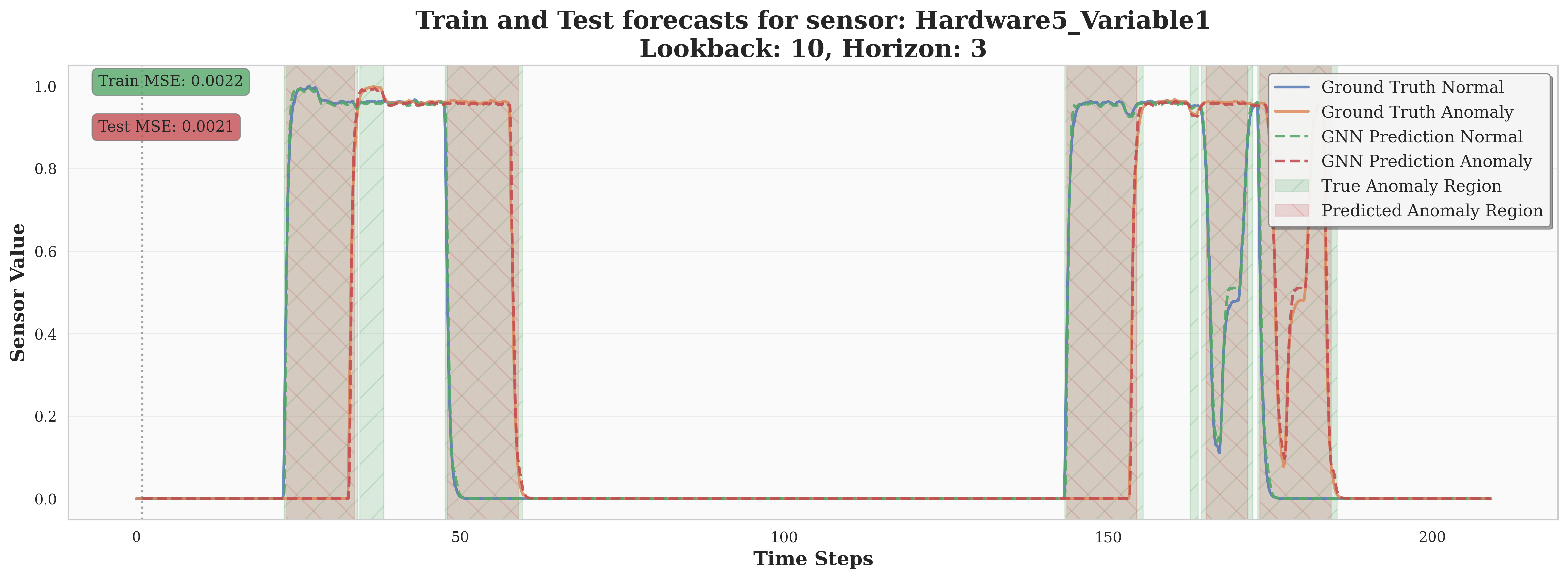}
    \end{subfigure}
    \vspace{0.4cm}
    \begin{subfigure}[b]{0.32\textwidth}
        \centering
        \includegraphics[width=\textwidth]{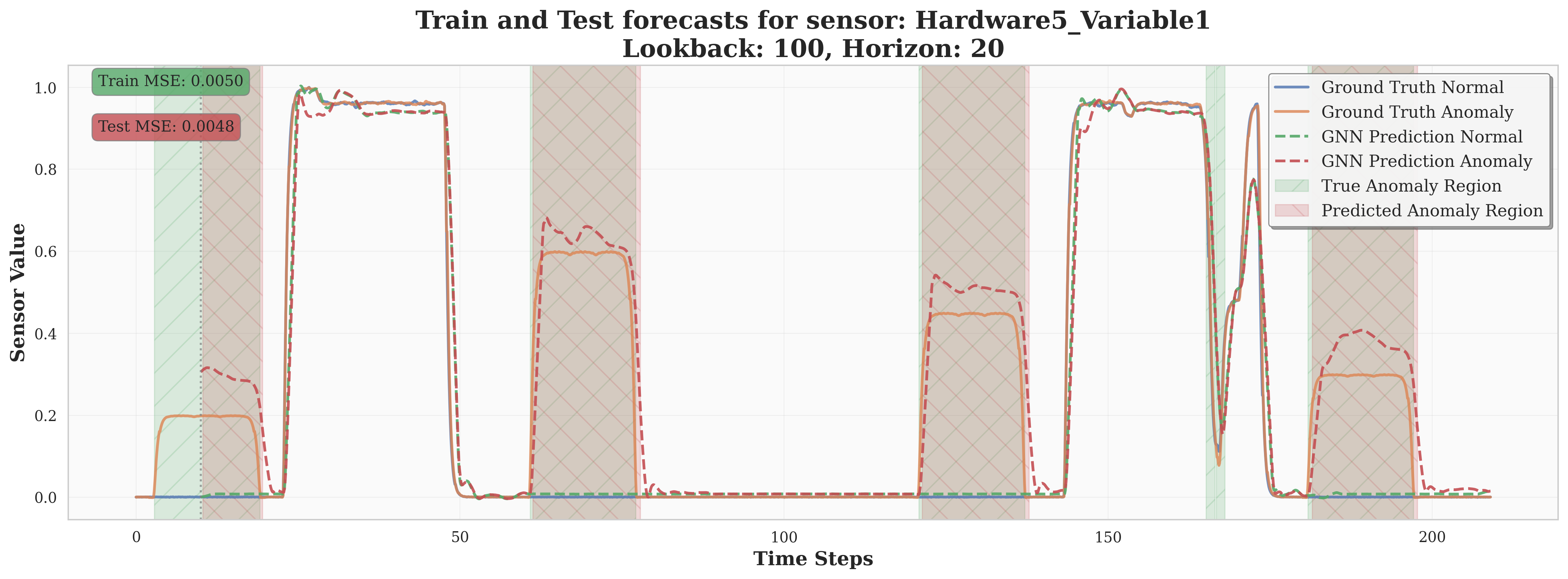}
        \caption{GNN forecasts for dataset A-1}
        \label{subfig:gnn4}
    \end{subfigure}
    \begin{subfigure}[b]{0.32\textwidth}
        \centering
        \includegraphics[width=\textwidth]{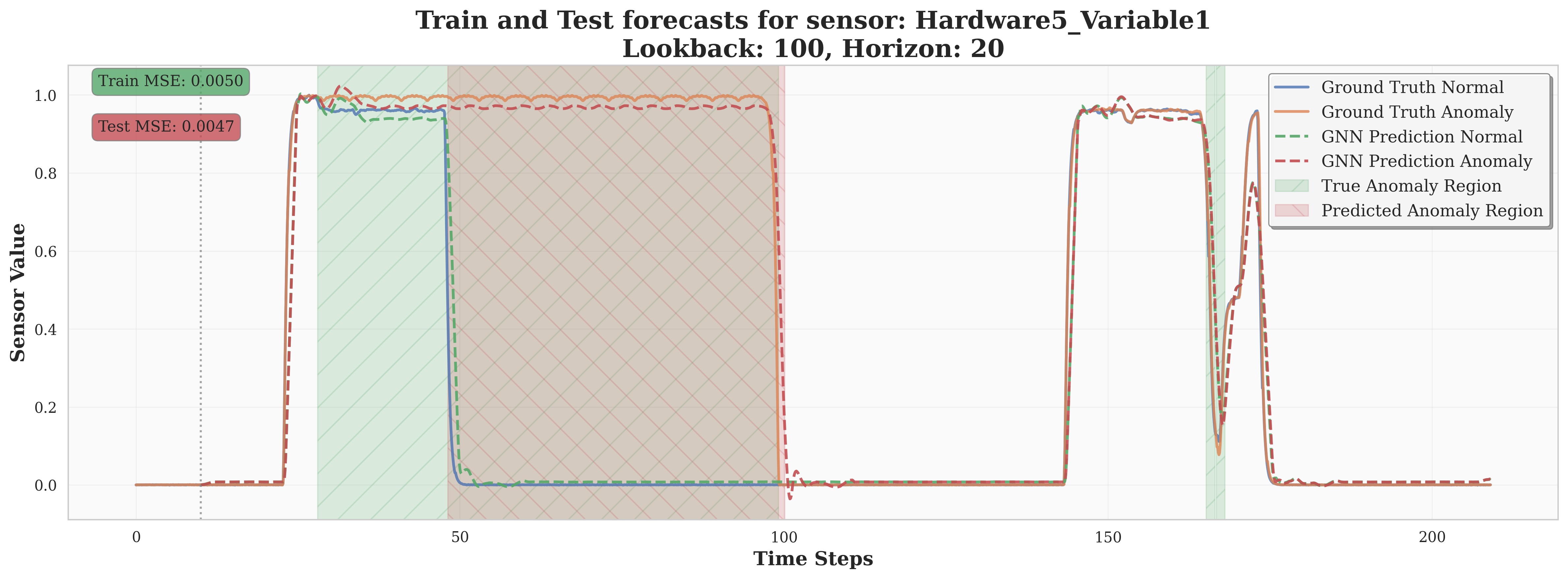}
        \caption{GNN forecasts for dataset A-3}
        \label{subfig:gnn5}
    \end{subfigure}
    \begin{subfigure}[b]{0.32\textwidth}
        \centering
        \includegraphics[width=\textwidth]{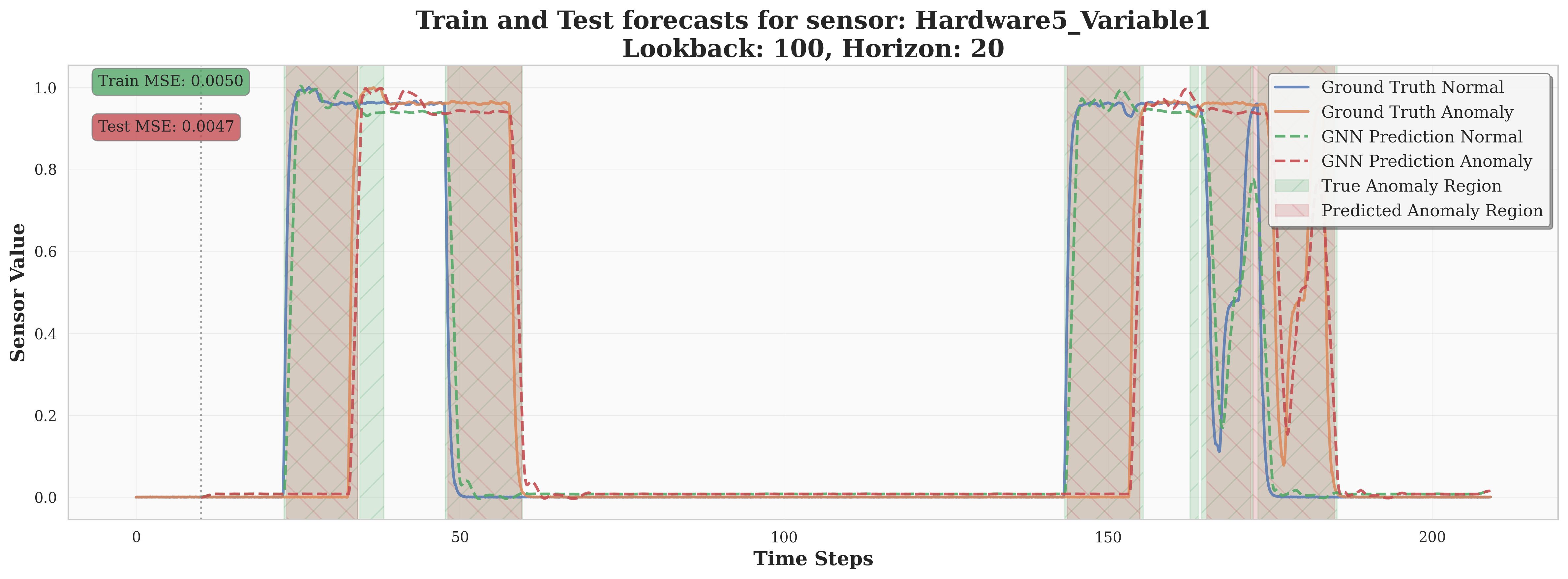}
        \caption{GNN forecasts for dataset A-9}
        \label{subfig:gnn6}
    \end{subfigure}
    \caption{Examples of forecasts by N-BEATS and GNN for the anomalous sensor in datasets A-1, A-3 and A-9, for two window configurations: (10,3) and (100,20). Blue and orange solid lines show the ground truth normal and anomalous signals, respectively. Green and red dashed lines display the corresponding forecast, and the green and red shaded regions mark the true and predicted anomalous time points.}
    \label{fig:NBeats_forecasting}
\end{figure*}

To illustrate the forecasting and anomaly detection capabilities, representative time series examples are provided in Figure \ref{fig:NBeats_forecasting}. As can be observed, both models have an excellent forecast performance for both the normal and anomalous signals for the (10,3) window configuration, leading to a near perfect detection of anomalous time points. When the window is increased to (100,20), both models' forecasting performance of the anomalous signal degrades while the normal signal forecasts is not impacted. The worse forecast of anomalous signals did not significantly impact the detection of anomalous time points which remained very accurate for both models. This diminished impact of forecast degradation on the anomaly detection was generally observed across datasets and reveals some robustness to the approach. In order to observe significant degradation in the anomaly detection metrics, the forecasting performance must first decrease significantly more. This occurs since it is sufficient for the forecast of the anomalous signal to deviate from the normal signal without necessarily reconstructing the anomaly accurately. Lastly, by comparing the forecast of anomalous signals by the two models, it is observable that the GNN's forecast performance degrades less than the N-BEATS', as was highlighted in Figs. \ref{fig:nbeats_result} and \ref{fig:gnn_result}.

    
    
    

\subsection{\textit{Top-K} ablation study}
\label{subsec:topk}
As explained in section \ref{subsubsec:MethGNN}, the \textit{top-K} parameter limits the number of neighbours per node, thereby reducing both the message-passing complexity and the computational cost of the attention mechanism. This parameter plays a crucial role in defining the graph structure and enables a controlled trade-off between model expressivity and efficiency. 

\begin{figure}
    \centering
    \includegraphics[width=0.9\linewidth]{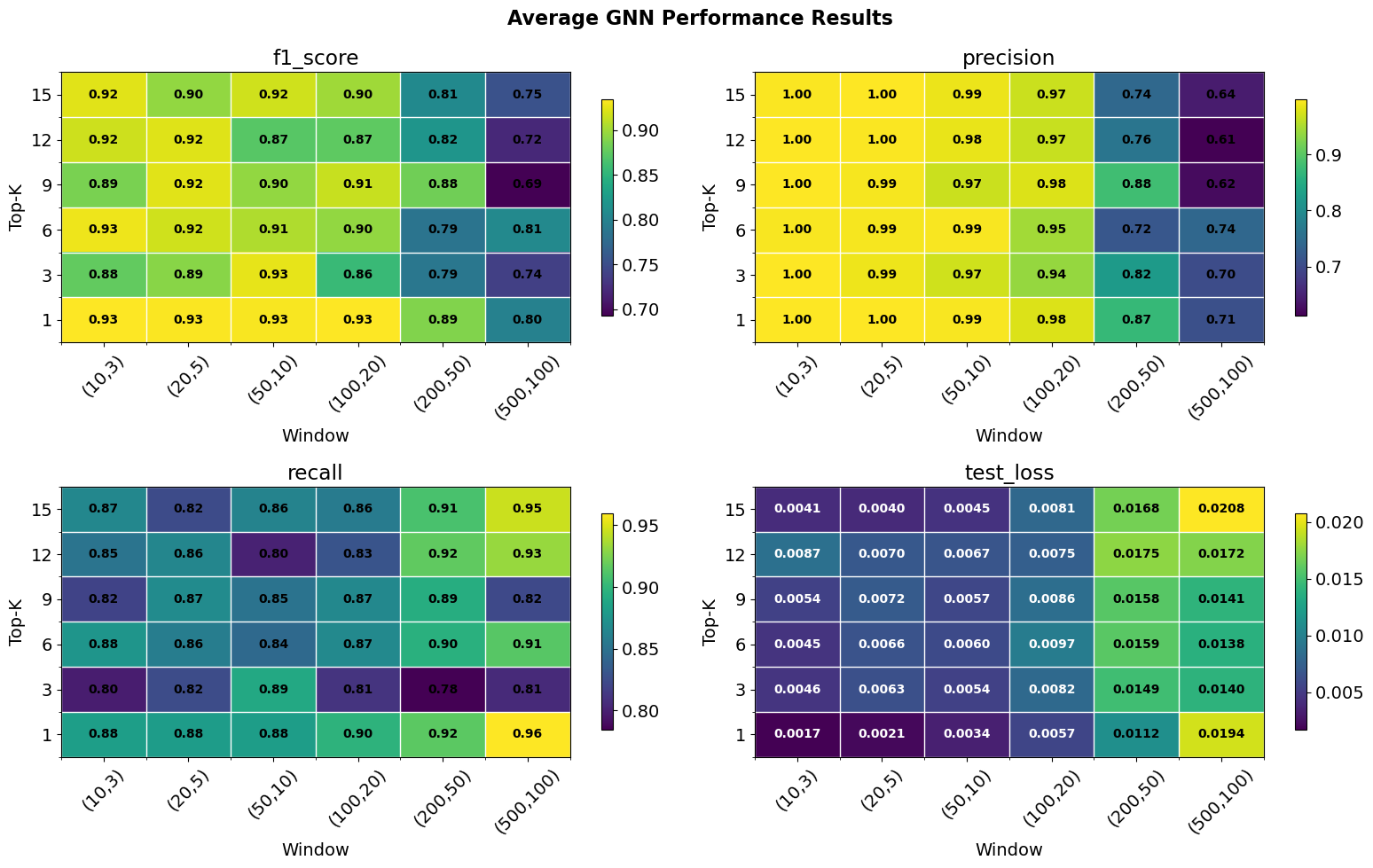}
    \caption{Ablation study results of \textit{top-K} parameter on GNN model performance.}
    \label{fig:Ablation}
\end{figure}

To assess the impact of \textit{top-K} on GNN performance, we trained separate models for each window configuration, each \textit{top-K} value listed in Table \ref{tab:hyperp}, and across all datasets. The results revealed consistent patterns across datasets; therefore, for conciseness, we present the average performance metrics in Fig. \ref{fig:Ablation}. 

From the figure, it is evident that the \textit{top-K} parameter has limited influence on anomaly detection performance. Detection accuracy appears to be primarily governed by the window configuration, remaining stable up to the (100,20) window before declining for larger horizons. However, when examining the test loss, \textit{top-K}=1 GNNs consistently achieve lower loss values, except in the largest window configuration (500,100).

\begin{figure}[htpb]
    \centering
    \begin{subfigure}[b]{0.48\linewidth}
        \centering\includegraphics[width=\linewidth]{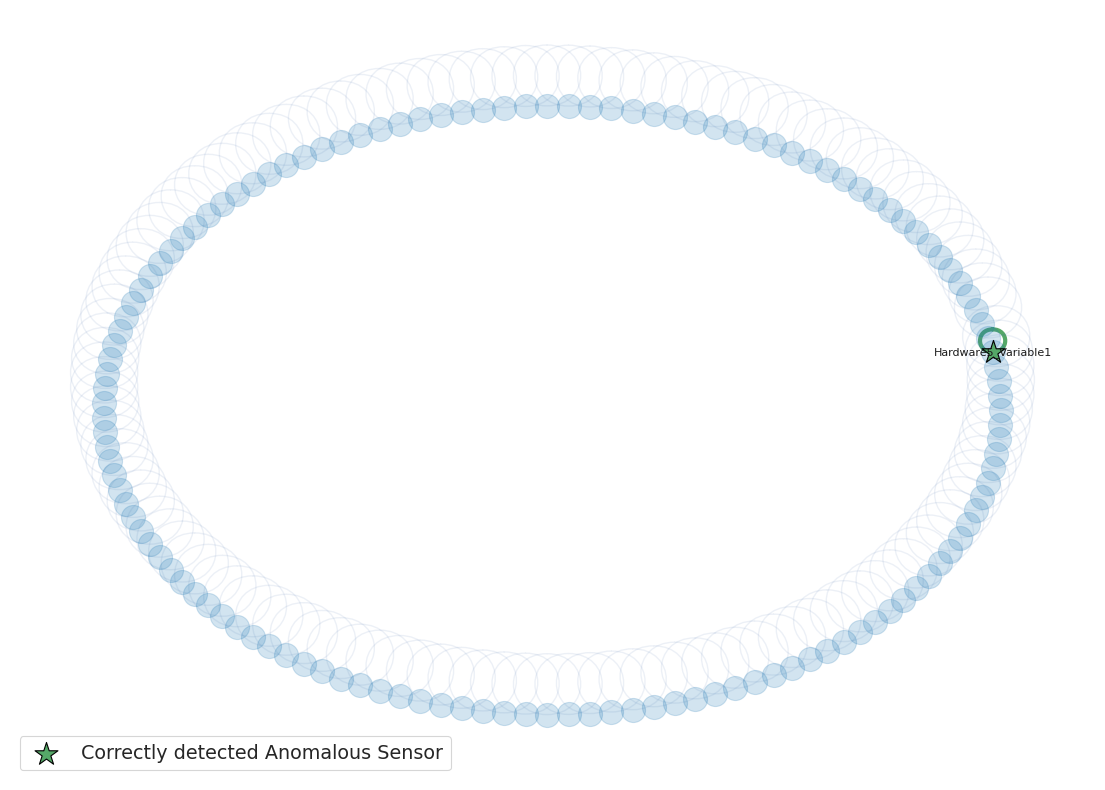}
        \caption{Exemplary GNN graph for \textit{top-K}=1}
        \label{subfig:Graph1}
    \end{subfigure}
    \vspace{0.4cm}
    \begin{subfigure}[b]{0.48\linewidth}
        \centering\includegraphics[width=\linewidth]{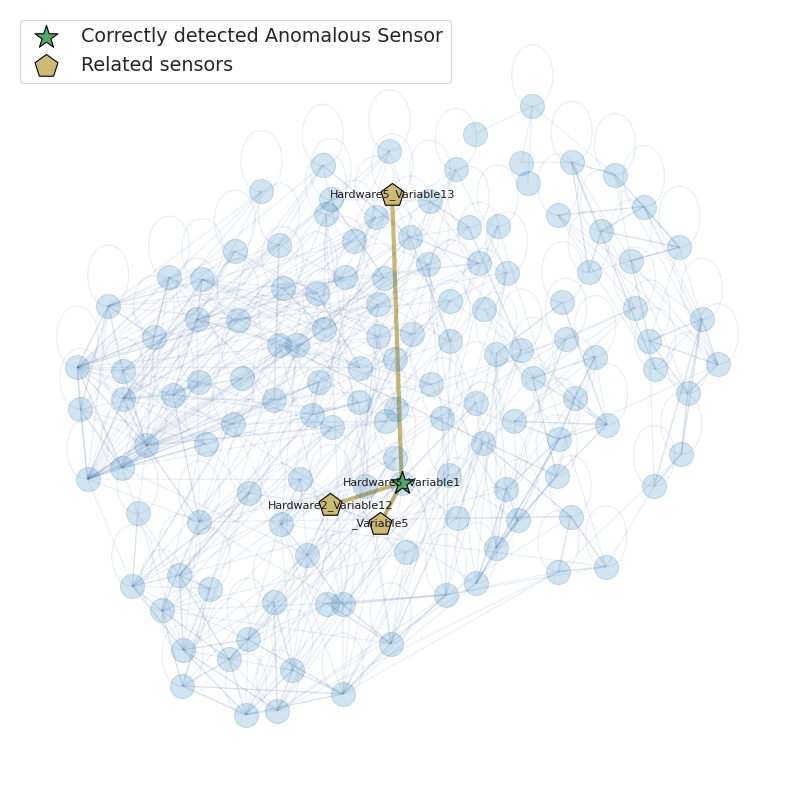}
        \caption{Exemplary GNN graph for \textit{top-K}=6}
        \label{subfig:Graph6}
    \end{subfigure}
    
    \caption{Exemplary Graphs from models applied to dataset A-1}
\end{figure}

This outcome is unexpected, as setting \textit{top-K}=1 severely restricts the multivariate nature of the model by allowing each node to receive messages from only one neighbour. In practice, this results in self-loops dominating the attention matrix, with nodes attending themselves only. This can be visualized in Figure \ref{subfig:Graph1}. Consequently, the model aggregates no cross-variable information. 

This behaviour suggests at least one of two explanations: (1) the GNN architecture is ineffective at leveraging cross-variable dependencies, or (2) the datasets used in this study contain minimal inter-variable relationships. According to the responsible process engineers, strong cross-variable dependencies were not expected, although their absence could not be definitely confirmed. As a result, it remains inconclusive why the model with \textit{top-K}=1 achieved the best performance.

A drawback of \textit{top-K}=1 displaying the best performance is the model's inability to find related sensors to the anomalous one. This capability was a key features of the model proposed in \cite{Deng2021}, as it allows for the detection of other sensors that might show anomalous behaviour. This limitation is of course only relevant in situation (1). 

\begin{table*}[ht]
\centering
\resizebox{\textwidth}{!}{%
\begin{tabular}{cc|ccccc|ccc|}
\multirow{2}{*}{Lookback} & \multirow{2}{*}{Horizon} & \multicolumn{5}{c|}{N-BEATS} & \multicolumn{3}{c|}{GNN} \\ \cline{3-10} 
 &  & \begin{tabular}[c]{@{}c@{}}Train time per\\ variable {[}mm:ss{]}\end{tabular} & \begin{tabular}[c]{@{}c@{}}Total train\\ time {[}hh:mm:ss{]}\end{tabular} & \begin{tabular}[c]{@{}c@{}}Test time per \\ variable {[}s{]}\end{tabular} & \begin{tabular}[c]{@{}c@{}}Total test \\ time {[}s{]}\end{tabular} & \begin{tabular}[c]{@{}c@{}}\#Trainable\\ Parameters\end{tabular} & \begin{tabular}[c]{@{}c@{}}Train\\ time {[}mm:ss{]}\end{tabular} & \begin{tabular}[c]{@{}c@{}}Test \\ time {[}s{]}\end{tabular} & \begin{tabular}[c]{@{}c@{}}\#Trainable\\ Parameters\end{tabular} \\ \hline
10 & 3 & 00:39 & 01:25:38 & 0.40 & 52.5 & 104,066 & 01:28 & 1.22 & 19,587 \\
20 & 5 & 00:35 & 01:15:32 & 0.39 & 51.5 & 106,746 & 01:26 & 1.49 & 21,125 \\
50 & 10 & 00:30 & 01:05:22 & 0.40 & 53.0 & 114,776 & 02:01 & 2.41 & 25,610 \\
100 & 20 & 00:39 & 01:26:00 & 0.41 & 53.1 & 128,176 & 02:00 & 3.75 & 33,300 \\
200 & 50 & 01:03 & 02:16:57 & 0.40 & 52.1 & 155,076 & 03:29 & 6.88 & 49,970 \\
500 & 100 & 01:17 & 02:48:39 & 0.39 & 51.5 & 235,376 & 06:17 & 11.45 & 94,820 \\ \hline
\end{tabular}%
}
\caption{Time and space complexity of the N-BEATS and GNN models trained and tested on the group-A datasets.}
\label{tab:complexity}
\end{table*}

\subsection{Computational complexity}
\label{subsec:complex}
In the previous section, we observed that the GNN achieves its best performance when \textit{top-K}=1, implying that the model does not leverage cross-variable information for its forecasts. This raises the question of whether a GNN architecture is justified for the task, as opposed to using N-BEATS model or another univariate model. However, the GNN remains a compelling choice due to its superior performance compared to N-BEATS, as demonstrated in section \ref{subsec:ResPerf}, and its significantly lower computational cost as now shown in Table \ref{tab:complexity}.

Table \ref{tab:complexity} highlights the disparity in computational complexity between the two models. For a single variable, the N-Beats requires between 2.5 and 5.3 times more trainable parameters than the GNN does for all variables. This difference translates into substantially longer training times: while the GNN completes training in a matter of minutes, N-BEATS requires between 1.5 and 3 hours. The GNN's ability to outperform N-BEATS despite having fewer parameters underscores its efficiency and suitability for the task.

Lastly, the test time reported in Table~\ref{tab:complexity} corresponds to inference over the full time series length (ranging from 1492 to 2080 samples). In a real-time anomaly forecasting scenario, only a single sample needs to be processed at each time step, therefore, both models are capable of producing forecasts within the defined horizon window, and thereby supporting real-time anomaly detection and intervention.

\section{Limitations and Future Work}
\label{Sec:Limit}
While our methodology and experiments showcase the strong performance and computational efficiency of the GNN model, several limitations and unexplored directions remain.  Addressing these will be essential for extending the applicability and robustness of the proposed approach. In particular, we highlight the following areas for future investigation:

\subsection{Anomaly Variety}
In the current study, we focus exclusively on synthetic anomalies applied to variables exhibiting step-like behaviour. To further validate the robustness of the model, it is necessary to evaluate its performance on anomalies affecting variables with different characteristics, such as smooth trends or high variance (noisy) signals. These types of anomalies may present distinct patterns and detection challenges compared to those studied here.

Moreover, testing the model in real anomaly occurrences is a critical next step. Although such anomalies are rare due to well-tuned manufacturing tools, and often go undetected when they occur, existing anomaly detection systems do occasionally flag events that could serve as test cases. Collaboration with process engineers responsible for tool monitoring will be essential to access and validate these real-world examples, enabling a more comprehensive evaluation of the model's practical utility.

\subsection{Variable input adaptability}
Semiconductor fabrication involves a complex sequence of tools, each monitoring a unique set of process variables. For practical deployment, it would be highly beneficial to develop a model capable of adapting to varying numbers and types of input sensors across different tools. While constructing a single model to handle all sensors across the entire process is computationally infeasible, a flexible architecture that can generalize across tools would significantly simplify implementation. Particularly, if retraining or fine-tuning the model for each model can be avoided.

For N-BEATS, adapting to different input sizes is straightforward, each variable can be modelled independently. However, as observed in section \ref{subsec:complex}, this approach incurs substantial computational overhead. In contrast, the GNN model faces challenges in adapting its sensor embedding matrix to accommodate new sensors without retraining. One potential solution, is to train a GNN on the largest sensor set available. and develop a dynamic routing mechanism that maps sensors from other tools to the trained graph based on behavioural similarity. Unused nodes could remain inactive during inference.

This approach requires much experimentation and validation but could lay the groundwork for a foundational time series model tailored to semiconductor manufacturing. Until such adaptability is achieved, the current model remains constrained to tool-specific applications and must be retrained before deployment.

\subsection{Extending to Causal Inference}
As discussed in section \ref{subsec:topk}, the GNN achieved its best performance without incorporating cross-variable information, raising questions about the presence of causal dependencies in the data. Importantly, the current model does not include any mechanism for explicitly detecting causal relationships; graph edges are constructed solely to minimize the forecasting loss. 

Incorporating causal inference could enhance both performance and interpretability. By identifying causal links prior to training, the graph structure could be informed by domain knowledge or data-driven causal discovery methods, such as those proposed by Liang \cite{Liang2021}. This would allow the GNN to focus its message passing on meaningful relationships, potentially improving forecasting accuracy and providing interpretable insights for process engineers.

Future work should explore integrating causal discovery algorithms into the graph construction phase, enabling the model to learn from and explain the underlying dynamics of the process more effectively.

\subsection{Predicting anomaly impact on device fabrication}
While the proposed models enable real-time anomaly forecasting, they do not provide insight into the downstream impact of these anomalies on device performance. Understanding this relationship is crucial for optimizing process parameters and ensuring product quality. Currently, process engineers rely on experience to infer which sensor anomalies affect specific device characteristics, but the scale and complexity of the process make this task challenging.

Establishing a connection between forecasted anomalies and device characterization would be highly valuable. One possible direction is to extend the GNN architecture to incorporate device-level outputs, effectively linking process anomalies to fabrication outcomes. Inspiration can be drawn from models such as NeuCube \cite{Kasabov2014}, which allow for the expansion of learned graphs by adding new nodes to extract higher-level information. This approach has shown promise in fields like neuroscience for modeling EEG data \cite{Doborjeh2018}.

\section{Conclusion}
\label{Sec:Conclusion}

In this study, we extended and applied two forecasting models to address the challenge of online anomaly prediction in semiconductor manufacturing. The models were trained to predict future time points based on non-anomalous trace data extracted from tool log-files, and anomalies were identified by comparing forecasts against anomalous instances. We explored both an univariate approach using N-BEATS, which assumes independence between sensors, and a multivariate approach using a Graph Neural Network (GNN), which captures inter-variable relationships through graph-based message passing.

Both models demonstrated strong forecasting accuracy up to a horizon of 20 time points and maintained reliable anomaly detection up to 50 time points. Notably, the GNN consistently outperformed N-BEATS across datasets, despite having between 2.5 and 5.3 times fewer trainable parameters. Surprisingly, the best GNN performance was achieved when the graph was limited to one edge per node—effectively resulting in a disconnected graph composed of self-loops. This finding raises questions about the necessity of cross-variable modeling in this context, yet the GNN's superior performance and computational efficiency justify its continued use.

Overall, we have developed a lightweight and effective model for online anomaly forecasting, suitable for deployment on manufacturing tools. Future work will focus on expanding the model's adaptability, interpretability, and integration with downstream device characterization to further enhance its utility in semiconductor process monitoring.

\section{Acknowledgement}
\label{Sec:Acknowledgement}

The authors would like to express their sincere gratitude to \textbf{Dr. Yasutoshi Okuno}, \textbf{Mr. Jun Kawai}, and \textbf{Mr. Hiroshi Horiguchi} from \textbf{SCREEN Holdings Co., Ltd.}, \textbf{SCREEN Advanced System Solutions Co., Ltd.}, and \textbf{SCREEN Semiconductor Solutions Co., Ltd.} for their valuable insights, constructive feedback, and engaging discussions, which significantly contributed to the direction, execution, and interpretation of this research work.


\end{document}